\pdfoutput=1

\documentclass[11pt]{article}

\usepackage{graphicx}
\usepackage{caption}  
\usepackage{subcaption} 
\usepackage{booktabs}
\usepackage{multirow}
\newcommand{\titletext}{Can VLMs Recall Factual Associations From Visual References?}
\newcommand{\entity}[1]{`\textsc{#1}'}
\usepackage{amsmath,amsfonts,bm}

\def\eqref#1{equation~\ref{#1}}

\def\1{\bm{1}}

\DeclareMathAlphabet{\mathsfit}{\encodingdefault}{\sfdefault}{m}{sl}
\SetMathAlphabet{\mathsfit}{bold}{\encodingdefault}{\sfdefault}{bx}{n}

\usepackage[preprint]{acl}

\usepackage{times}
\usepackage{latexsym}

\usepackage[T1]{fontenc}

\usepackage[utf8]{inputenc}

\usepackage{microtype}

\usepackage{inconsolata}

\usepackage{graphicx}

\title{\titletext}

\author{Dhananjay Ashok\textsuperscript{$\ddagger$}, Ashutosh Chaubey\textsuperscript{$\dagger$}, Hirona J. Arai\textsuperscript{$\dagger$}, Jonathan May\textsuperscript{$\ddagger$} and Jesse Thomason\textsuperscript{$\dagger$} \\
  \textsuperscript{$\dagger$} University of Southern California \\
  \textsuperscript{$\ddagger$}  Information Sciences Institute, University of Southern California \\
  \texttt{\{ashokd, jonmay\}}@isi.edu, \texttt{\{achaubey, hjarai, jessetho\}}@usc.edu
  }

\begin{document}
\maketitle
\begin{abstract}
Through a controlled study, we identify a systematic deficiency in the multimodal grounding of Vision Language Models (VLMs).
While VLMs can recall factual associations when provided a textual reference to an entity, their ability to do so is significantly diminished when the reference is visual instead. 
Forcing VLMs to rely on image representations of an entity halves their ability to recall factual knowledge, suggesting that VLMs struggle to link their internal knowledge of an entity with its image representation. 
We show that such linking failures are correlated with the expression of distinct patterns in model internal states, and that probes on these internal states achieve over 92\% accuracy at flagging cases where the VLM response is unreliable. 
These probes can be applied, without retraining, to identify when a VLM will fail to correctly answer a question that requires an understanding of multimodal input. 
When used to facilitate selective prediction on a visual question answering task, the probes increase coverage by 7.87\% (absolute) while also reducing the risk of error by 0.9\% (absolute).  
Addressing the systematic, detectable deficiency is an important avenue in language grounding, and we provide informed recommendations for future directions. 

\begin{figure}[th]
    \centering
    \includegraphics[width=0.9\linewidth]{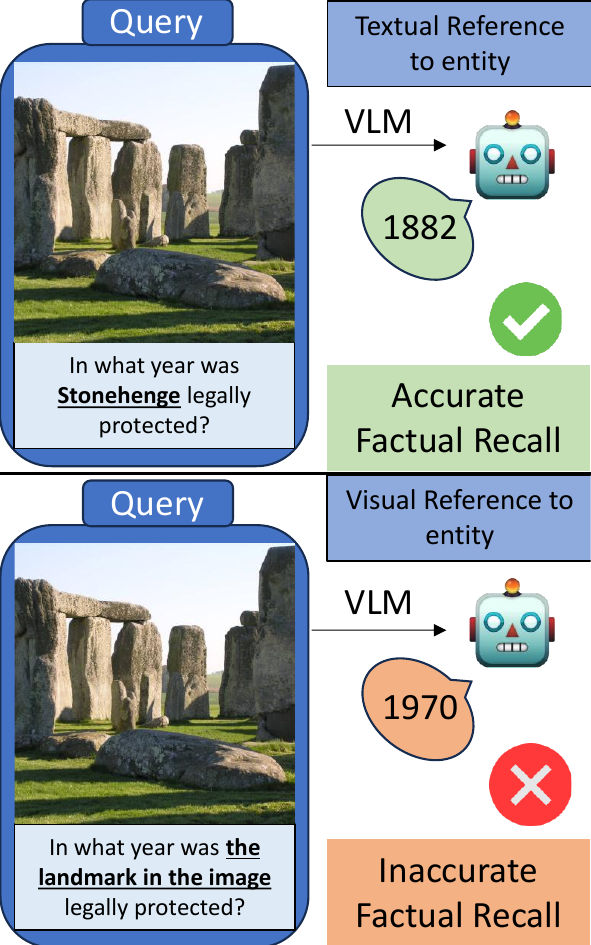}
    \caption{Despite being able to recall its internal knowledge of an entity (here, \entity{Stonehenge}) when provided a textual reference, LLaVa-Vicuna-7B~\citep{liu2023visual} fails to recall this knowledge with a visual reference.}
    \label{fig:teaser}
\end{figure}

\end{abstract}

\section{Introduction}
\label{sec:introduction}

\begin{figure*}[th]
    \centering
    \includegraphics[width=1\linewidth]{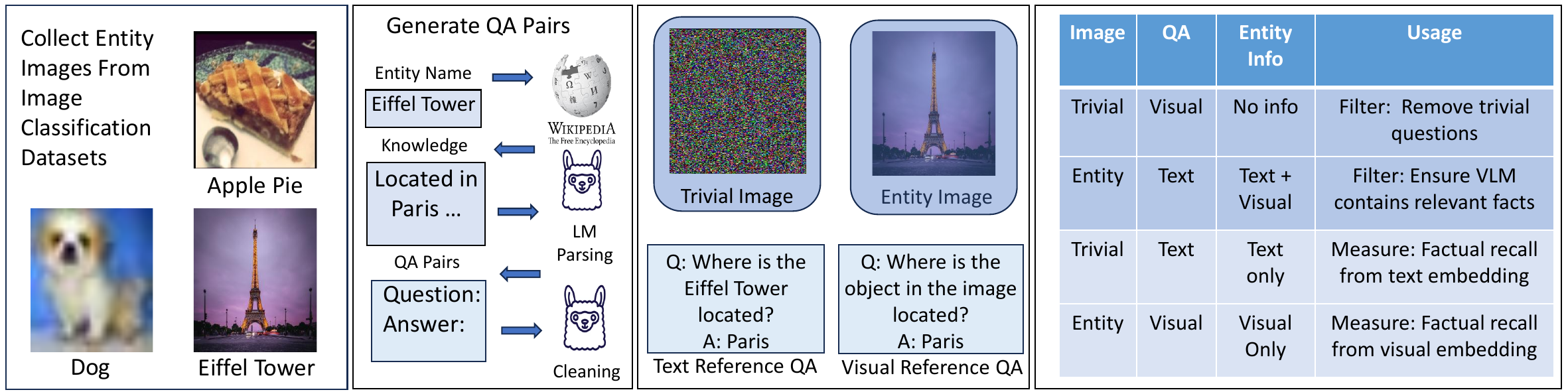}
    \caption{
        For each question, we create a \textit{textual reference} version which names the entity in the text, and a \textit{visual reference} version in which the text only refers to the image. Similarly, we have an entity image that identifies the entity, and an uninformative image (e.g., uniformly random pixels) which contains no information about the entity. 
        We precisely control the modality that contains entity-specific information, isolating the ability of a VLM to recall the factual knowledge it contains based on the modality in which this information is presented.}
    \label{fig:mainfig}
\end{figure*}

Systems that incorporate multiple modalities, such as image and text, address the `symbol grounding problem,' i.e., the problem of connecting symbolic concepts to sensory properties of objects in the world~\citep{harnad1990symbol}. It has been argued that addressing this problem is a prerequisite to truly `understanding meaning'~\citep{bender2020climbing}, making multimodal grounding a vital frontier in the pursuit of capable AI systems~\citep{bisk2020experience, bommasani2021opportunities}. 

An important instance of this problem is that of vision-language grounding~\citep{kazemzadeh2014referitgame}, with modern Vision Language Models (VLMs) making significant progress on tasks that combine the visual and textual modalities~\citep{antol2015vqa, alayrac2022flamingo}. VLMs contain facts on a wide range of entities~\citep{petroni2019language}, and can use this knowledge to reason over objects present in an image~\citep{zellers2019recognition}.

A consistently grounded multimodal system should recall facts about a symbolic entity (e.g., \entity{Stonehenge}) independent of the sensory modality by which it is observed (e.g., the word `Stonehenge' versus an image of the same, Figure~\ref{fig:teaser}).  

In this work, we test several state-of-the-art VLMs for this capability. We design a controlled experiment to isolate their ability to recall facts when using visual representations of the entity, and observe, on average, a 58.95\% performance degradation in visual question answering capability.

We evaluate seven VLMs across the InstructBLIP~\citep{li2023blip}, LLaVA~\citep{liu2023visual} and GPT4-V~\citep{gpt4vsystem} families, encompassing a variety of vision-language pretraining paradigms, backbone language models and model scales.\footnote{\url{https://anonymous.4open.science/r/VLM_Grounding/}}
Every VLM we investigate, without exception, experiences significant performance degradation when forced to rely on its image representation, with answer accuracy consistently falling by over 50\%. The decline in performance is persistent across models of varying sizes, suggesting that scaling the model is insufficient to fully address this grounding gap. 
VLMs struggle to link their internal knowledge of entities with their image representation, relying heavily on the presence of text tokens that explicitly name the entity and failing to achieve the modality independence expected by a consistently grounded model.

Having established that VLMs often fail to link images of entities to their internal knowledge, we next turn to detecting and avoiding such failures. We employ techniques of mechanistic interpretability~\citep{logitlens} on the hidden states of the VLM and visualize how predictions of the next output token are built during inference. We observe a consistent pattern: cases of linking failure build confidence in a VLM's prediction at later layers when compared to cases of linking success. 
We use linear probes on these hidden states to create warning systems for such linking failures. The probes significantly outperform the next best baseline (by 25 percentage points on average), achieve accuracies in excess of 92\% and generalize to out-of-domain datasets. These probes can be used, without retraining, to determine whether or not a VLM will correctly answer a question from the OKVQA~\citep{marino2019ok} dataset. When used in a selective prediction~\citep{de2000reject} framework, our method achieves higher coverage than all baselines, with an absolute improvement of 7.87\%, while also reducing errors by 0.9\%.

\section{Background and Related Work}
\label{sec:relatedwork}
Efforts to jointly model textual and visual modalities have made significant progress in recent years~\citep{zhang2024vision}, growing from approaches that were limited to representing isolated word meanings~\citep{barnard2005word} to powerful systems that can handle arbitrary image and textual inputs on a wide variety of tasks~\citep{alayrac2022flamingo, li2023blip}. 

These methods take advantage of powerful pretrained language models (LMs)~\citep{zhang2024vision}, training visual encoders that convert images to representations that can be passed into the LMs~\citep{liu2023visual}, with minimal changes to the LM itself~\citep{li2023blip}. Since pretraining LMs on internet-scale corpora imbues them with factual knowledge~\citep{petroni2019language}, the resulting system can recall facts about entities present in the images they are provided~\citep{marino2019ok}. 

This approach to vision-language grounding does not jointly learn representations, raising concerns of a grounding gap between the visual representations and the LM component~\citep{li2023evaluating, Tong_2024_CVPR}. Qualitative studies on the semantic representations of such VLMs reveal notable differences between multimodal and text-only representations of words~\citep{pezzelle-etal-2021-word, tikhonov2023leverage}. However, whether such differences affect downstream performance, such as the capability of a VLM to access the internal knowledge that it contains, remains an open question.

Prior work evaluates VLMs on their ability to recall factual knowledge by asking knowledge-intensive visual questions~\citep{antol2015vqa,marino2019ok,Cheng2025SimpleVQAMF, das-etal-2024-exams, saikh2022scienceqa}. However, these approaches are oriented towards benchmarking the overall capability of VLMs, and are not able to isolate failures in multimodal grounding specifically. A failure to answer a question from these benchmarks could be due to a failure to link the visual representation with the VLMs internal knowledge, however, it may also be due to a failure in identifying the image, a gap in the knowledge contained in the VLM, a reasoning error, or some combination of these. We create an experiment that controls for such confounding factors, isolating the ability of a VLM to access its internal knowledge using only visual representations of an entity. While contemporary work explores this problem~\citep{cohen-etal-2025-performance} from the lens of mechanistic interpretability, they focus only on the PopVQA dataset. In this work, we construct a testbed from multiple datasets that span a range of entities, showing that VLMs struggle to recall factual associations from visual references over a wide variety of entity types.


\section{Creating the Benchmark}
\label{sec:benchmark}
We start by collecting images of various entities from image classification datasets, with each class label serving as a distinct entity. We source from:

\noindent{\textbf{CIFAR100}~\citep{krizhevsky2009learning}:} A general-purpose dataset with entities ranging from insects such as \entity{beetle} to vehicles such as \entity{train}.

\noindent{\textbf{Food101}~\citep{bossard14}:} A food dataset, with entities including \entity{baklava} and \entity{ceviche}.

\noindent{\textbf{Landmarks}~\citep{weyand2020google}:} A dataset of famous landmarks, with entities such as \entity{Niagara Falls} and \entity{Stonehenge}.

For each entity, we compile general knowledge question-answer (QA) pairs.  Unlike most existing benchmarks~\citep{antol2015vqa, balanced_vqa_v2}, our questions are well-formed in text alone, i.e., the answer can be determined without referring to any specific, contextualized image of the entity. 

We use the Wikipedia API~\citep{wikiapi} to automatically gather text containing common knowledge about each entity, and use Llama3.1-8B~\citep{grattafiori2024llama} as a parser that extracts question-answer (QA) pairs from the text. We additionally directly prompt Llama3.1-8B to generate QA pairs for each entity. The collected pairs are filtered through multiple steps to ensure they are valid questions with a single correct answer (Appendix~\ref{sec:appendix_pipeline}). Finally, three authors of the paper conduct a round of human annotation and establish, with substantial inter-annotator agreement ($\kappa>0.65$), that the remaining QA pairs are both relevant to the entity and ones where the answer to the question is correct. 

For each question, we create an analogous version that does not state the entity but instead refers to the image. Ultimately, each datapoint in our testbed (Figure~\ref{fig:mainfig}) consists of an entity's image, a \textit{textual reference} question that explicitly mentions the entity and a \textit{visual reference} question that does not name the entity. Separate from this pipeline, we also design a task based on the MNIST~\citep{deng2012mnist} dataset. For each digit in the dataset, we randomly generate addition and multiplication questions with a single other operand of no more than 2 digits (e.g. for the digit \entity{5}, we may generate the \textit{textual reference question} `5 + 17 =' and \textit{visual reference question} `the digit in the image + 17 ='). 
 
 However, when given the image and \textit{visual reference} question, there are multiple reasons why a VLM may answer incorrectly. The VLM might have misidentified the entity, or may not contain the relevant information, which are scenarios where even a perfectly grounded system will fail. To control for these factors, we conduct an additional round of filtering. For each datapoint, we ask the VLM under evaluation to identify the object in the entity image. We also have it answer the question when given both the entity image and the \textit{text reference} version of the question. We retain the question only if the identification is accurate and the answer is correct, ensuring that the VLM recognizes the entity and contains the relevant internal knowledge. We also ask the VLM to answer the question when provided a trivial image with no information about the entity (e.g., random pixel values) and the \textit{visual reference} version of the question. These instances should be unanswerable as they do not contain the information required to identify the entity, and a correct response indicates that the question's answer can be guessed using language-based priors. We remove such instances to ensure that our testbed contains questions which require knowledge of the entity to answer correctly. In practice, we aggregate the VLMs output over multiple different trivial images, for details see Appendix~\ref{sec:appendix_vlm_inference}. 
The final benchmark allows us to isolate the VLMs ability to perform factual recall from different modalities in a way that is not possible with existing complex VQA benchmarks. Consider the visual reasoning question answering benchmark, GQA~\citep{hudson2019gqa}. This benchmark includes highly compositional images and may contain, for instance, an image of an apple and an orange on a table. We may create the \textit{textual reference question} `What colour is an apple?', which has the analogous \textit{visual reference question} `What color is the object in the image?'. However, since the image contains multiple objects, the correct answer to the textual reference question (red) is no longer the only correct answer to the visual question (red or orange). By sourcing our images from simple, single-image datasets, we can precisely isolate the entity and ensure that the textual and visual questions have the same answers. 

The final number of datapoints varies by dataset and model, with an average of \textbf{792} distinct image-question pairs remaining per dataset-model combination. For dataset-model specific counts, see Appendix~\ref{sec:appendix_pipeline}

\begin{figure*}[t]
    \centering 

    \begin{subfigure}[t]{0.49\textwidth} 
        \centering
        \includegraphics[width=0.99\linewidth]{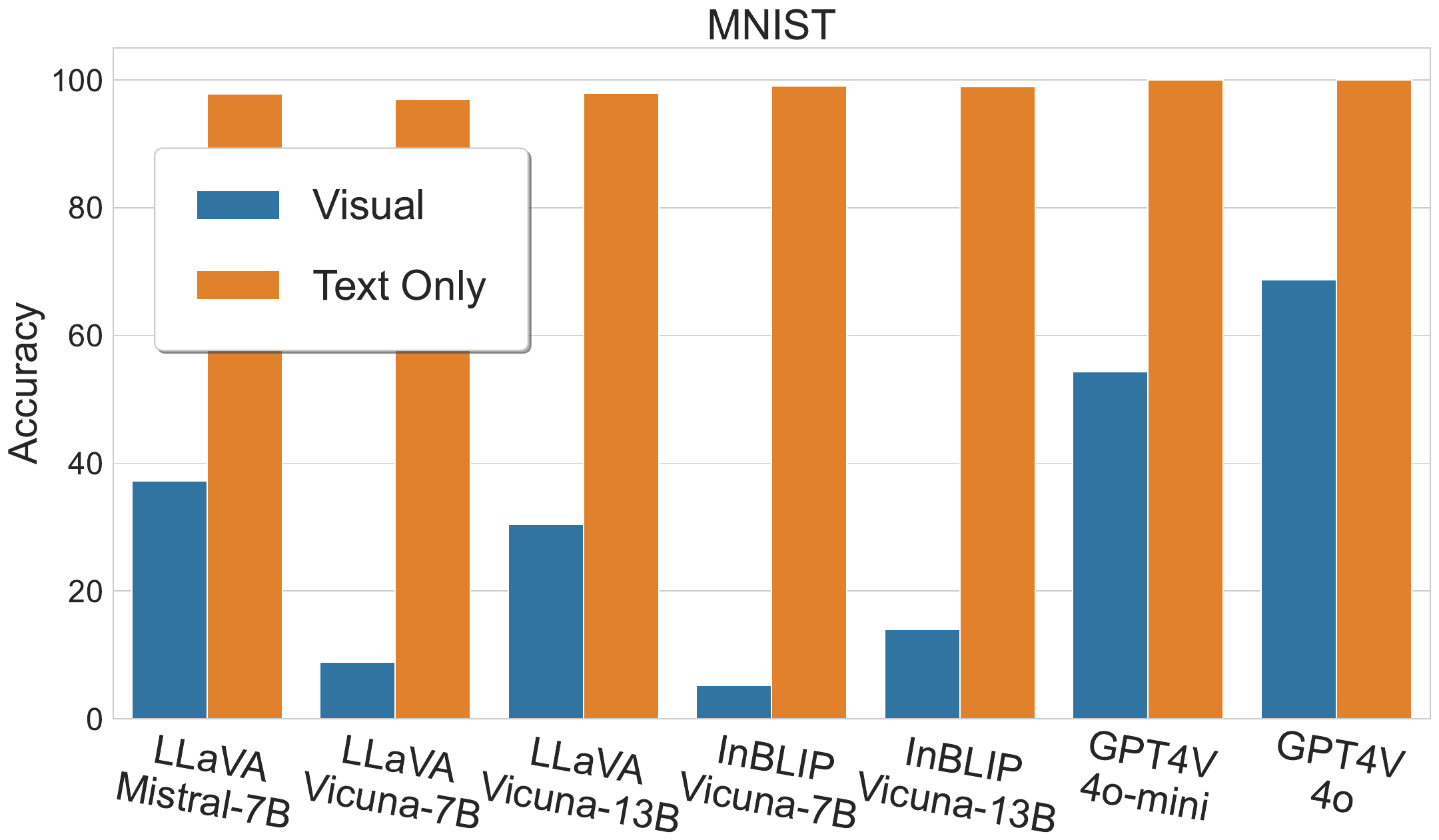}
    \end{subfigure}
    \hspace{0.1cm}
    \begin{subfigure}[t]{0.49\textwidth}
        \centering
        \includegraphics[width=0.99\linewidth]{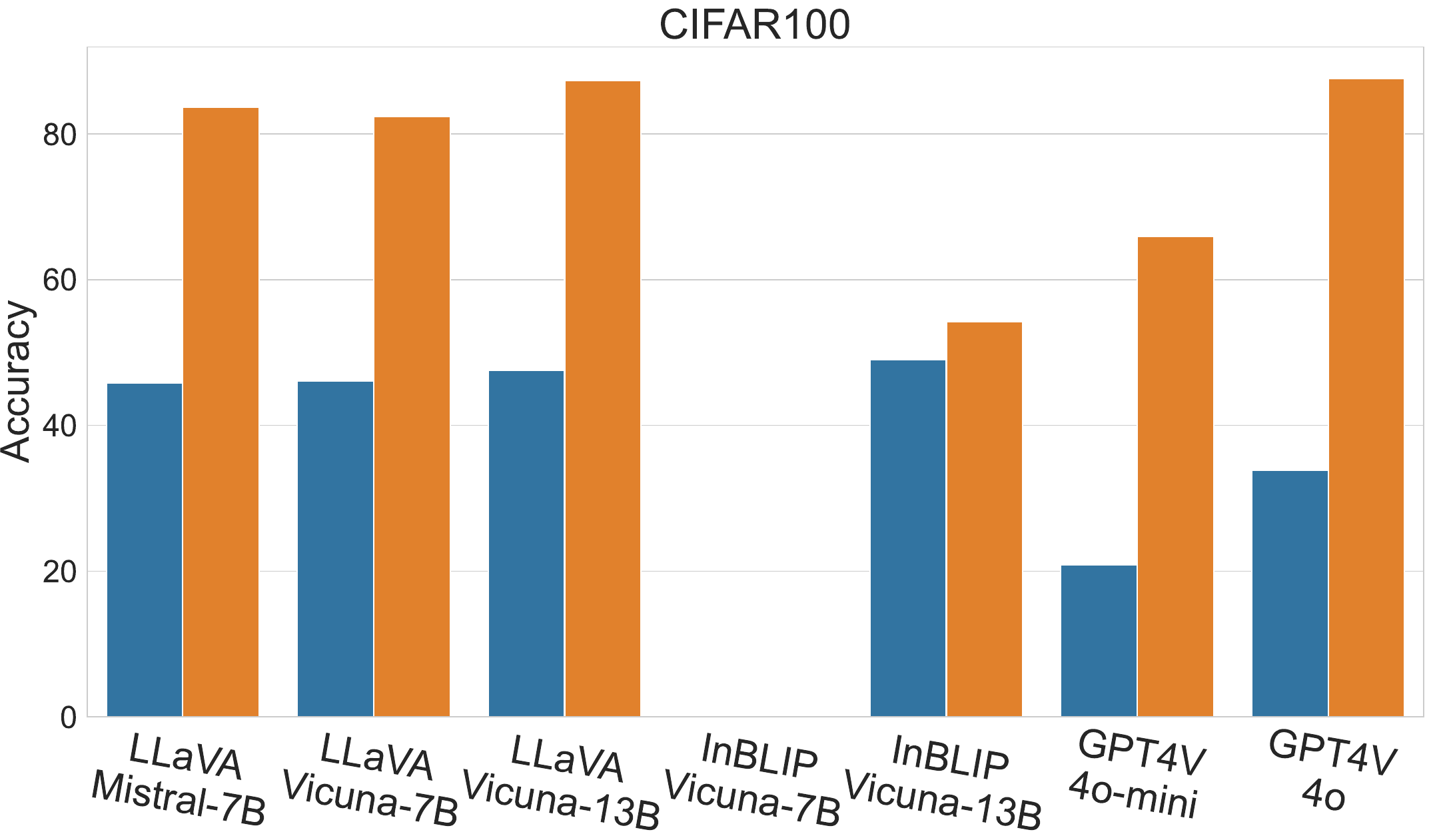}
    \end{subfigure}%

    \vspace{0.5cm} 

    \begin{subfigure}[t]{0.49\textwidth}
        \centering
        \includegraphics[width=0.99\linewidth]{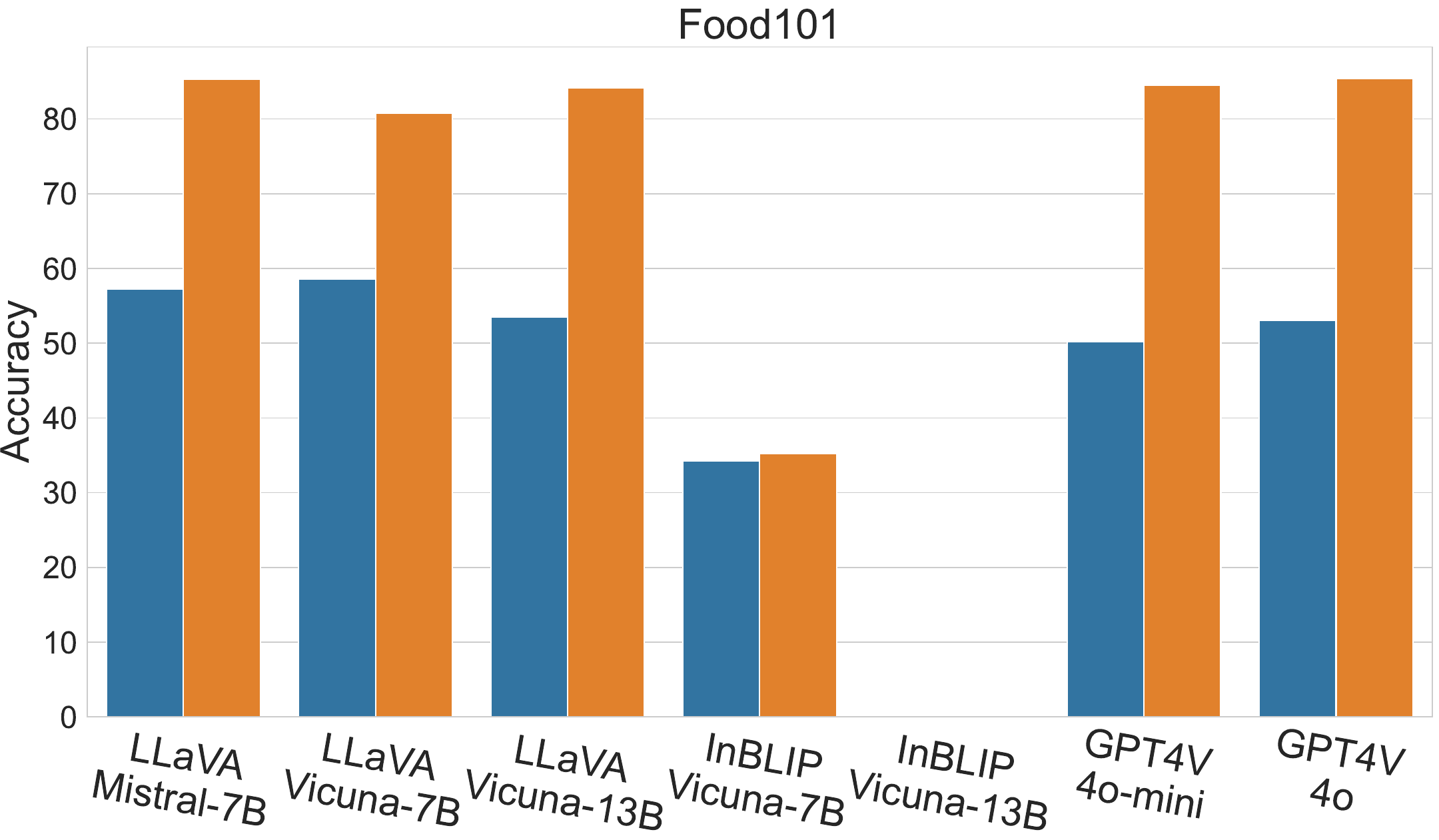}
    \end{subfigure}%
    \hspace{0.1cm} 
    \begin{subfigure}[t]{0.49\textwidth}
        \centering
        \includegraphics[width=0.99\linewidth]{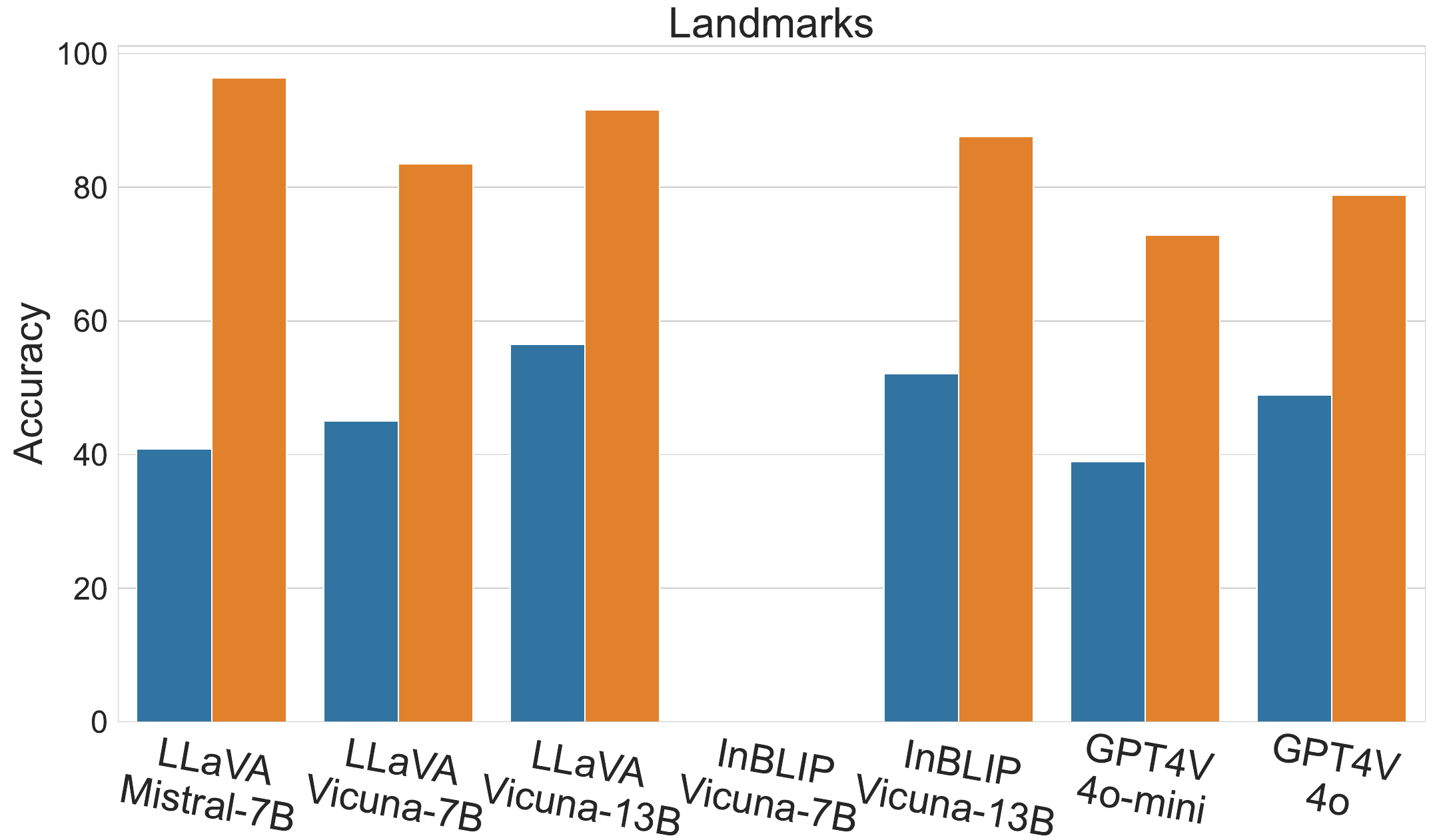}
    \end{subfigure}%
    \caption{Performance (two-way inclusion) on VLMs when forced to rely on visual (\textcolor{blue}{Visual}) or textual (\textcolor{orange}{Text Only}) representations of an entity. Empty bars for InstructBLIP are datasets where fewer than 100 datapoints passed filters. While the \textcolor{orange}{Text Only} setting shows performance decreases (15.3\%), the performance in the \textcolor{blue}{Visual} declines by 58.95\% (nearly 4x the decline). This decline is evidence that VLMs, across pretraining paradigms, architectures and scales, struggle to link their internal knowledge of an entity with visual representations of it.}
    \label{fig:experiment}

\end{figure*}

\begin{figure*}[t]
    \centering 

    \begin{subfigure}[t]{0.49\textwidth} 
        \centering
        \includegraphics[width=0.8\linewidth]{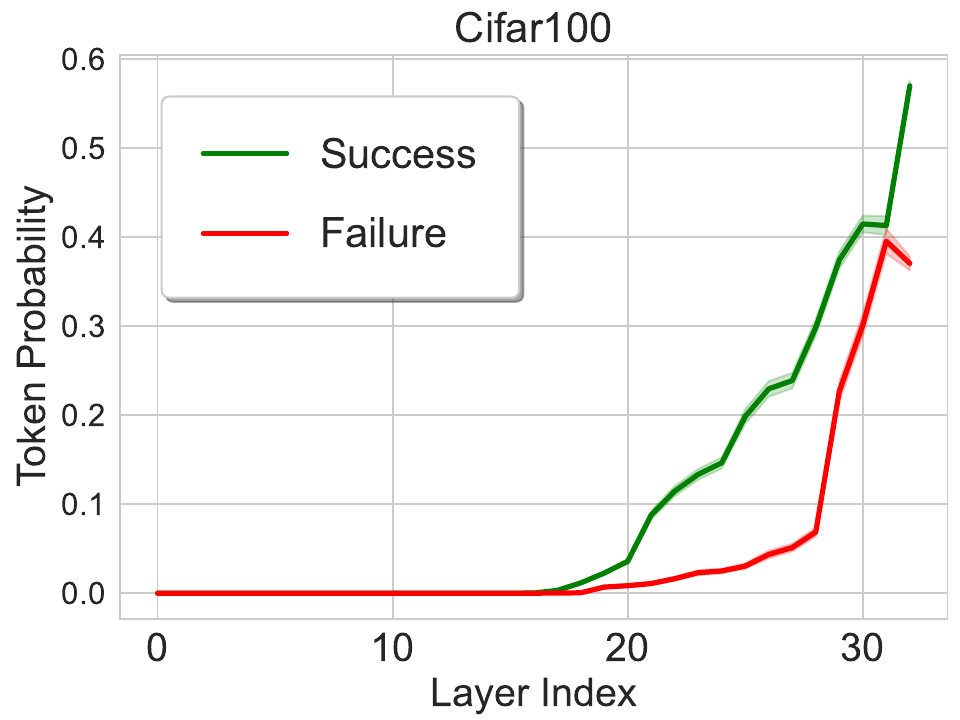}
    \end{subfigure}
    \hspace{0.01cm}
    \begin{subfigure}[t]{0.49\textwidth}
        \centering
        \includegraphics[width=0.8\linewidth]{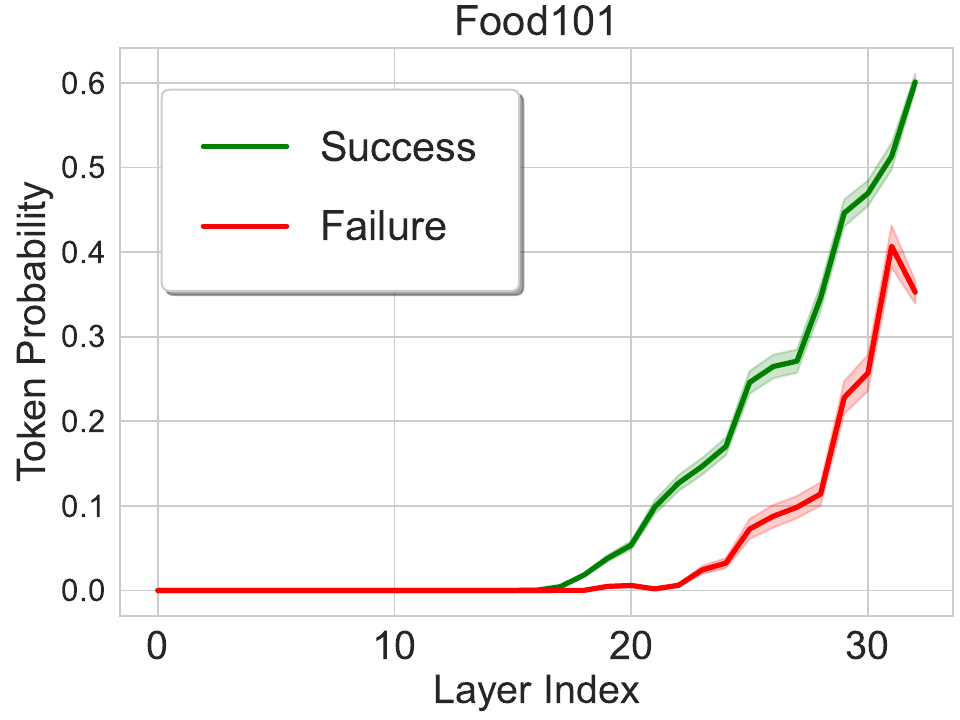}
    \end{subfigure}%

    \vspace{0.5cm} 

    \hrule

    \begin{subfigure}[t]{0.49\textwidth}
        \centering
        \includegraphics[width=0.8\linewidth]{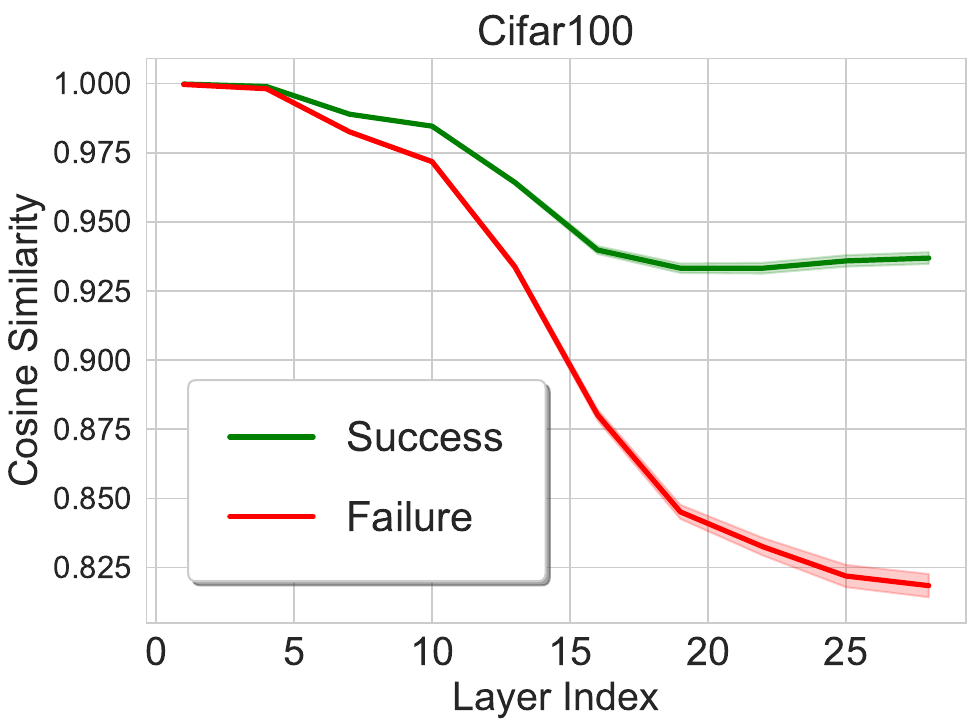}
    \end{subfigure}%
    \hspace{0.01cm} 
    \begin{subfigure}[t]{0.49\textwidth}
        \centering
        \includegraphics[width=0.8\linewidth]{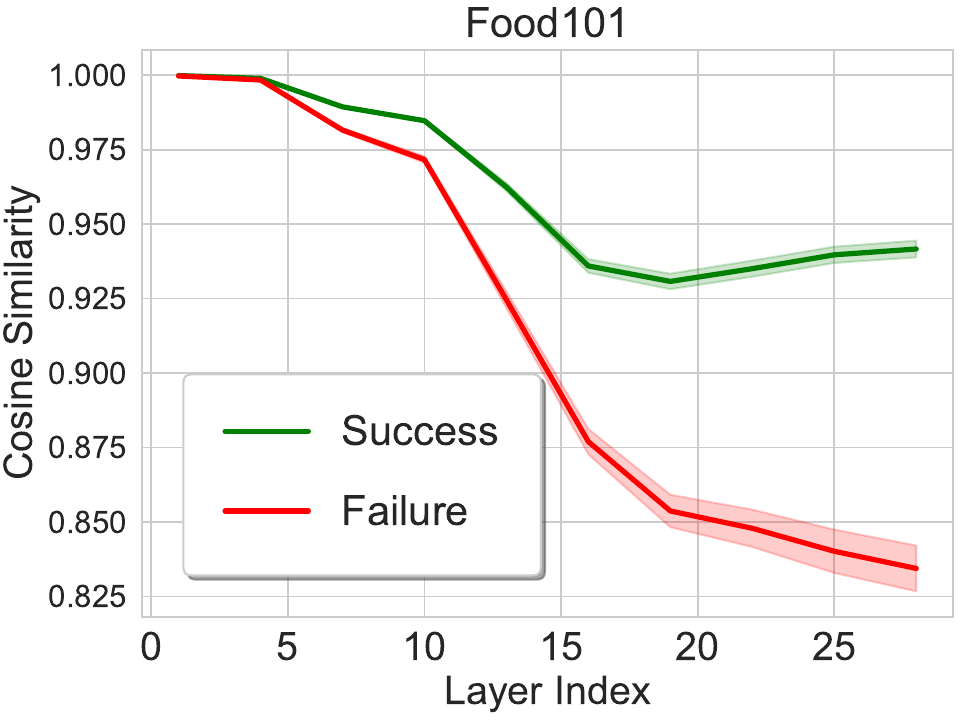}
    \end{subfigure}%
    \caption{Cases of linking failure correlate with the expression of distinct patterns in the hidden states produced by the VLM. \textbf{Top:} 
 The probability of the eventual output token across layers of the forward pass. Cases of linking success gain probability mass earlier than linking failures, and achieve higher eventual probability on average. \textbf{Bottom:} The Cosine Similarity between internal states produced by the VLM when run in the \textcolor{blue}{Visual} and  \textcolor{magenta}{Full Info} settings (in the \textcolor{magenta}{Full Info} setting, the VLM is given both textual and visual entity references). The internal states of linking failure cases are less similar to the internal states produced in the \textcolor{magenta}{Full Info} setting, suggesting that linking failure occurs when the internal representations lack the information required to trigger the recall of relevant facts.}
    \label{fig:analysis}

\end{figure*}

\section{Evaluating VLMs Factual Recall using Different Modalities}
We evaluate the VLMs under two different settings, isolating the capability of the LM component to use the textual and visual representations to recall the relevant fact regarding an entity. In both settings, the VLM is prompted to directly state the answer.

\noindent\textcolor{orange}{\textbf{Text Only:}} We provide a trivial image and the \textit{textual reference question} that states the entity. This setting measures the VLM's ability to recall factual associations from the textual representation of the entity. 

\noindent\textcolor{blue}{\textbf{Visual:}} We provide the image of the entity, and the \textit{visual reference} question that does not name the entity in text. This setting forces the VLM to rely on its representation of the image to identify and access the facts it has stored about the entity.




We evaluate VLMs from the following families:

\noindent\textbf{InstructBLIP:}  An approach that leaves the LM and vision encoder frozen, learning a Q-Former~\citep{li2023blip} to convert image representations to visual tokens that the LM can use. 

\noindent\textbf{LLaVA:} A method that keeps the vision encoder frozen, but learns a linear projection from the image representation to the input space of the LM and also learns updates to the LM weights. 

\noindent\textbf{GPT4V:} A state-of-the-art frontier model whose internal details are proprietary. 

To evaluate a model response string $r$ against the gold answer string $a$ we use the metrics of two-way string inclusion ($r \in a \lor a\in r$, where $\in$ is the substring property) and exact match ($a=r$), which estimate the correctness of the response with a binary signal. We also use BLEU~\citep{papineni2002bleu}, a continuous metric which gives higher values when the answer and response have greater n-gram overlaps. 

\section{Results}
\label{sec:results}
A consistently grounded VLM that can access its internal knowledge using references from either modality should achieve comparable performance in both test settings. The results (Figure~\ref{fig:experiment}) show that answer correctness under the \textcolor{orange}{\textbf{Text Only}} setting, as judged by the two-way inclusion metric, is consistently high at 84.70\% on average. This performance strikes a stark contrast with the average correctness of the \textcolor{blue}{\textbf{Visual}} setting (42.05\%), which is less than half of the performance of the \textcolor{orange}{\textbf{Text Only}} setting. This divide is also seen when using the exact match (52.87\% vs 23.85\%) and BLEU (55.93\% vs 21.92\%) metrics (Appendix~\ref{sec:appendix_metric}). When recalling factual associations regarding an entity, VLM performance declines by over half if it is forced to rely on visual representations of the entity instead of textual representations.

The grounding gap, which occurs consistently across all models and datasets, is a systematic failure in the ability of VLMs to recall factual associations using image representations of entities. This gap is most chronic in the MNIST dataset, where the \textcolor{orange}{\textbf{Text Only}} setting achieves near perfect performance ($98.66\%$), but the \textcolor{blue}{\textbf{Visual}} setting achieves only $31.28\%$ correctness.

Concerningly, the gap continues to be wide even for the most powerful VLM, GPT4V, suggesting that it is a failure that will not be resolved naturally through a scaling of model parameters and dataset sizes. Within the same model family, models of higher scales occasionally have less of a gap; however, as seen in the Food101 and Landmarks datasets, the improvement is often marginal.

\begin{table*}[t]
\begin{tabular}{@{}lrrrrrrrr@{}}
\toprule
\multicolumn{1}{c}{Dataset} & \multicolumn{2}{c}{Perplexity}                          & \multicolumn{2}{c}{Probe}                               & \multicolumn{2}{c}{Ensemble}                            & \multicolumn{2}{c}{$\Delta$ (Perplexity$\to$Ensemble)}                                         \\ 
\cmidrule(lr){2-3} \cmidrule(lr){4-5} \cmidrule(lr){6-7} \cmidrule(l){8-9} 
                            & \multicolumn{1}{c}{Coverage} & \multicolumn{1}{c}{Risk} & \multicolumn{1}{c}{Coverage} & \multicolumn{1}{c}{Risk} & \multicolumn{1}{c}{Coverage} & \multicolumn{1}{c}{Risk} & \multicolumn{1}{c}{Coverage (\%)} & \multicolumn{1}{c}{Risk (\%)} \\ \midrule
CIFAR100                    & 43.52                        & 22.86                    & 52.41                        & 23.22                    & 52.12                        & 22.33                    & +19.76                       & -2.31                  \\
Food101                     & 56.78                        & 15.85                    & 59.18                        & 16.83                    & 60.38                        & 15.90                     & +6.34                       & +0.31                  \\
Landmarks                   & 51.92                        & 29.42                    & 54.28                        & 31.12                    & 53.78                        & 29.33                    & +3.58                       & -0.30                  \\
OKVQA                       & 57.97                        & 42.08                    & 66.58                        & 39.23                    & 59.02                        & 41.53                    & +1.81                       & -1.30                  \\ \bottomrule
\end{tabular}
\caption{Performance of probes when used for selective prediction in the out-of-distribution setting. Probes are trained on all three other datasets and applied to the test dataset without retraining. Coverage (percentage of datapoints the system predicts on) is typically higher for the probes, however the perplexity baseline can incur lower risk (error-rate on predicted samples). An ensemble approach that combines the two often leads to higher coverage than the perplexity baseline with a minor decrease in the error-risk.}
\label{table:okvqa}
\end{table*}

\section{Identifying Linking Failures}
Our controlled experiments have shown significant declines in the ability of a VLM to access its internal knowledge of an entity using visual references, a grounding gap which makes these systems more vulnerable to false utterances. Such responses can be particularly harmful in settings where users depend on the model~\citep{macleod2017understanding}, trust in its output~\citep{ji2023survey} or use its output to make consequential decisions~\citep{xie2024large}. 
In this section, we develop methods that identify when a VLM will fail to link its visual representation of an entity with its internal knowledge, allowing model providers to flag instances where the response should not be trusted, and minimize the harms of linking failures. 
We focus on LLaVA-Vicuna7B and investigate the hidden representations of the last input token in the LM component. During the forward pass on datapoint $i$, specifically after the computation at layer $l$ of a Transformer LM layer, the last input token is represented by a $d$-dimensional vector $h^{(l)}_{i}$. 

\subsection{Visualizing Hidden States}
The unembedding layer of the VLM is a linear operation (of form: $U\in\mathbb{R}^{|V|, d}$) trained to map the last token embedding of the penultimate layer to the vocabulary space. 
We use this layer to compute $\text{softmax}(Uh_i^{(l)})$, which can be seen~\citep{logitlens, geva2021transformer} as an approximation of the distribution of the next token over the vocabulary space at a given layer of the VLM. We study how this distribution varies for the successful and unsuccessful cases of linking the visual reference of an entity with internal knowledge about that entity. However, the very first token generated by the VLM may be the same for both correct and incorrect answers, preventing us from observing a clear trend. Hence, we use Llama-3.1-8B to reformulate the QA pairs into multiple-choice QAs with four options (prompts in Appendix~\ref{sec:appendix_pipeline}). Taking inspiration from the setup in \citet{rimsky-etal-2024-steering}, the first token generated is now the letter associated with the option that the VLM selects, and hence completely determines whether or not the question has been answered correctly. Figure~\ref{fig:analysis} (Top) shows how the model builds its prediction over the forward pass. Cases of linking success begin gaining probability earlier than linking failures, and achieve higher eventual probability on average. The patterns suggest that the layers most responsible for promoting the final answer vary based on whether or not factual linking is successful. The layers responsible for factual answers are the mid-to-late layers, 15-25. Answers that gain most of their probability mass after this range are often a result of linking failure. The special role of the mid-to-late layers in our results is consistent with previous work ~\citep{meng2022locating, azaria2023internal}, which suggests that these layers play a disproportionate role in the storage of factual associations. 

We also compare the differences between the hidden states generated when the modality of entity information varies. Specifically, for a given datapoint, we run in both the \textcolor{blue}{Visual} and \textcolor{magenta}{Full Info} settings (the \textcolor{magenta}{Full Info} setting provides the VLM with both the entity image and the \textit{textual reference} to the entity). These two runs produce the hidden states $\{h_{i}^l|l\in\{1, 2, \ldots L\}\}_{i=1}^N$ and $\{\tilde{h}_{i}^l|l\in\{1, 2, \ldots L\}\}_{i=1}^N$ respectively (for an LM with $L$ layers). To perform a layerwise comparison of the internal states produced by both of these runs, we compute the cosine similarity of $h_{i}^{(l)}$ and $\tilde{h}_{i}^{(l)}$ for every datapoint, at every layer. 

Unlike the probability distributions, the cosine similarities of the successful and failing linking cases diverge steadily (Figure~\ref{fig:analysis}, Bottom). By the final layers (in the \textit{visual} setting), the internal states of the successful cases of linking are substantially more similar to the internal states produced in the \textit{full information} setting. These patterns suggest that linking fails when the internal state does not contain sufficient information or a strong enough signal to trigger the recollection of relevant facts.

\subsection{Early Warning for Linking Failure}
A failure to link an entity representation with the VLMs internal knowledge could lead to hallucination and ungrounded outputs. A system that can notify users when such linking failures have occured, would better allow them to calibrate their belief in the VLMs response. We build such a system, training a probe to flag cases when a VLM has failed to link its internal knowledge of an entity with its visual representation. We collect the 20th layer, final output token hidden states produced during the \textit{visual} setting and train a linear probe~\citep{alain2017understanding} to identify whether or not linking has failed. Similar approaches have been used to identify false utterances~\citep{azaria2023internal}, and a range of other degenerate behaviours in LMs~\citep{ashok2025language}. We consider the base rate of the most frequently occurring class to be the random baseline performance. For a stronger baseline, we implement a dominant method in LM confidence elicitation~\citep{kumar2024confidence} and selective prediction~\citep{srinivasan2024selective} with VLMs---learning a threshold on the per-token-perplexity of the output, and flagging for linking failure if the answer token perplexity lies above this threshold. Our linking detection probes (Table~\ref{table:probe}) significantly outperform all baselines. 

\begin{table}[ht]
\begin{tabular}{@{}lrrr@{}}
\toprule
\multicolumn{1}{c}{Dataset} & \multicolumn{1}{c}{Random} & \multicolumn{1}{c}{Perplexity} & \multicolumn{1}{c}{Probe} \\ \midrule
MNIST                       & 81.46                      & 54.97                          & 99.34                     \\
CIFAR100                    & 64.01                      & 64.98                          & 98.07                     \\
Food101                     & 66.47                      & 67.66                          & 94.61                     \\
Landmarks                   & 52.80                       & 68.32                          & 92.55                     \\
OKVQA                       &  51.34                             &   60.16                             &      64.32                       \\
\bottomrule
\end{tabular}
\caption{On all datasets, our probes significantly outperform all baselines, showing that the information contained in the hidden states allows us to build a powerful predictor for linking failures. }
\label{table:probe}
\end{table}

Next, we ask whether these probes can generalize to unseen datasets and prove useful in identifying linking failures under shifting distributions. We train a single probe on the CIFAR100, Food101 and Landmark datasets and apply this probe, without retraining, to the OKVQA dataset~\citep{marino2019ok}. This dataset is a testbed for commonsense knowledge in visual question answering, and asks for knowledge related to the entities in the image. We use the signal from our probe to enable selective prediction on OKVQA, i.e., deciding when the VLM should abstain from answering a question because it is unlikely to answer correctly. 

Following previous work~\citep{srinivasan2024selective}, we evaluate selective prediction with coverage and risk~\citep{el2010foundations}. Given a labelled evaluation dataset with image, question, and answer tuples, the coverage is the percentage of questions where the system makes a prediction, and the risk is the error rate on the questions where a prediction is made. 

The probes achieve (Table~\ref{table:okvqa}) higher coverage than the perplexity baseline with marginally higher risk. However, when we average the predictions of the two methods (Ensemble), we can achieve significantly higher coverage (10.4\% on average) than the perplexity baseline while simultaneously reducing the risk (-0.9\%). This strong performance demonstrates the value of hidden state probes in practical settings and their ability to identify linking failures under shifting distributions.

\section{Addressing the Grounding Gap}

The controlled experiments above have demonstrated that VLMs struggle to link their internal knowledge of an entity with a visual representation of it. However, one might argue that the VLM can always verbalize the image's contents first, allowing it to rely on the textual representation for factual recall. We advance several arguments to motivate the claim that the grounding gap shown in this work will limit the performance of VLMs in practical settings. 

\noindent\textbf{Inference Cost:} The most direct consequence of relying on the VLM to verbalize the contents of an image before answering is an increase in the number of tokens generated before the final answer can be provided. With the increasing scale of LMs~\citep{hoffmanntraining}, this threatens to become prohibitively expensive for both users~\citep{zhou2024survey} and the environment~\citep{strubell2020energy}. Addressing this grounding gap would enable the visual representation of the entity to recall the information directly and enable more efficient and accessible systems. 

\noindent\textbf{Conflicts between text and image modalities:} In practice, the information in the text used to describe an associated image may conflict with the information in the image's contents. Recent work shows~\citep{parcalabescu2025do} that answer generation is primarily driven by the text component and that in such cases where the modalities conflict, text information is given more weight. A superior system would make contextual decisions based on the strength of the evidence in each modality. However, VLMs that rely disproportionately on the textual modality for factual recall will consistently favour the text modality and lack this capability. 

\noindent\textbf{Verbalization Failures:}  VLMs that rely on verbalization to recall factual associations face a failure mode---inability to access factual, internally stored information about the objects in the image, because they were not explicitly verbalized when describing the image. This concern is most pressing for images that have many objects or entities that the query could involve, leading the VLM to only state a subset of them, or hallucinate objects that are not in the image~\citep{li2023evaluating}. Certain VLM architectures, such as the LLaVA approach~\citep{liu2023visual}, create visual representations that are agnostic to the text of the input prompt. These approaches may be more vulnerable to such failures, because the image encoding used to drive the verbalization cannot allow for a greater focus on the regions of the image most relevant to the text query. 

\noindent\textbf{Lack of Visual Knowledge in Text Training Data:} LMs are trained on internet-scaled text corpora~\citep{brown2020language} and hence store factual associations from the texts they have encountered~\citep{petroni2019language}. However, the concepts that are well learned from text data often do not fully include visual concepts such as spatial understanding~\citep{yamadaevaluating}, leading to LMs that struggle on questions requiring spatial knowledge~\citep{wang2024picture}. If VLMs remain reliant on their LM subcomponents for factual recall, they risk being limited in their ability to answer questions requiring a deep understanding of such visual concepts. 


\noindent\textbf{A Novel Data Augmentation Approach:} We suggest a novel training paradigm, a supplement to current pretraining procedures that may help address the grounding gap shown in this work. While our results (Figure~\ref{fig:experiment}) show that VLMs struggle to recall factual associations from visual references, most models still retain some capacity to do this. This suggests that the various architectures of these models do not necessarily prevent them from being consistently grounded across modalities. We suggest a variation of the feature-alignment pretraining procedure used to construct the LLaVa models~\citep{liu2023visual}. The suggested procedure assumes only that the vision component and language component are linked with a trainable bridging module. 

The current practice is to use an image captioning dataset, where the inputs to the VLMs are images and the target output is the caption. The bridging module is tuned to maximize the likelihood of the caption given the image, hence serving as an interface between the visual and textual modalities. We suggest that this paradigm be expanded to include factual recall. This would include collecting QA pairs for the entities in the captions of the images, a procedure which, as shown in this paper, can be done automatically at moderate scales. Instead of only using the captioning task for feature alignment, the pretraining phase can include tuning the bridging module to maximize the likelihood of the answer given the question. This paradigm may allow the bridging module to learn how to transform the visual representation such that it can draw out relevant facts from the layers of the LM. 

\section{Conclusion}
In this work, we show that Vision Language Models of various pretraining paradigms, architectures and scales suffer from a multimodal grounding gap. Specifically, we use a controlled experiment to demonstrate that VLMs struggle to recall factual associations regarding an entity from a visual reference, as opposed to a textual reference. This failure to link an entity's visual representation with its internal knowledge correlates with the expression of distinct patterns in the activations of the hidden states, and probes trained on these hidden states show promise in identifying these cases as well as providing a signal for selective prediction. We argue that addressing this gap is an important avenue for future research and suggest novel future directions to do so. 

\section{Limitations}
\label{sec:limitations}
This work primarily focuses on robustly showing that VLMs struggle to link their internal knowledge of an entity with their visual representations. In our work, we focus exclusively on VLMs, leaving an open question as to whether or not similar gaps can be found in other multimodal LMs (e.g., audio language models). We provide tools to identify when a VLM has failed to link an entity with its internal knowledge and also suggest directions to bridge this gap. However, the paper does not address mitigating linking failure during VLM training. Additionally, we do not identify which particular design decision (training stage, data distribution, architectural decisions, etc.) leads to this grounding gap. However, our results show that models which have a wide variety of such decisions all exhibit this gap, suggesting that it is not a flaw that is introduced during the training process, but rather a capability that is never fully learned during training. We hope future work will deepen our understanding of this grounding gap and help build models that can robustly access their internal factual knowledge regardless of the modality of access. 

\section{Ethical Considerations}
\label{sec:ethical}
Persistent grounding gaps between the vision and textual modality make VLMs prone to object hallucinations~\citep{li2023evaluating} and diminish the robustness of these systems. In safety-critical settings, such hallucinations could cause active harm to individuals, making work that closes this grounding gap an important avenue of future research. However, a successful effort to close the grounding gap between modalities would make multimodal models more capable, which is not without its own risks~\citep{zhang2024multitrust}. We refer the reader to recent surveys on the potential harm of strong, multimodal foundation models~\citep{xu2025mmdt, li2024multimodal} for a description of such risks.

\section*{Acknowledgements}
Dhananjay Ashok and Jonathan May acknowledge support from Open Philanthropy. This research is supported by the U.S. Defense Advanced Research Projects Agency (DARPA) Other Transaction awards HR00112490374 and HR00112490376 from the Friction for Accountability in Conversational Transactions (FACT) program.
Any opinions, findings, conclusions, or recommendations expressed here are those of the authors and
do not necessarily reflect the view of the sponsors.



\bibliography{main}

\appendix
\section{Additional Details on QA Generation Pipeline}
\label{sec:appendix_pipeline}

The pipeline for QA data generation consists of the following steps:
\begin{enumerate}
    \item Retrieve the Wikipedia entry for a given entity
    \item Parse the entry to split it such that every split of the entry consists of no more than two sentences
    \item Remove all splits that do not contain the entity name
    \item For each split, prompt Llama-3.1-8B to generate question-answer pairs from it
\end{enumerate}

We note that this use case is in line with the license for the API~\url{https://www.mediawiki.org/wiki/Special:Version/License/MediaWiki}.

The questions then undergo multiple rounds of data cleaning:
\begin{enumerate}
    \item Remove pair if answer has more than 7 words
    \item Remove pair if answer contains entity. This is because we want to measure factual recall, not the ability to state the object in the image. 
    \item Remove pair if the question does not contain the entity. This is to ensure the text alone can identify the entity. 
    \item Remove pair if Llama-3.1-8B decides that its answer is not unique. This helps reduce the number of ambiguous, unclear or subjective questions that are not fact-based.  
    \item Remove pair if Llama-3.1-8B answers the question incorrectly. 
\end{enumerate}

Finally, the QA pairs are deduplicated using both exact match and Llama-3.1-8B. We then randomly sample 5 images of the entity and pair them with a single QA pair to make 5 datapoints in our testbed. The four image classification datasets (MNIST, Food101, Landmarks and CIFAR100) have either a CreativeCommons or GNU General Public License, making this use case permissible. 

The next set of filters are specific to each VLM. 

\begin{enumerate}
    \item Remove an Image, QA pair if the VLM cannot identify the entity in the image
    \item Remove an Image, QA pair if the VLM cannot answer the QA pair correctly when provided both the \textit{textual reference} question and the image (see Figure~\ref{fig:mainfig} and Section~\ref{sec:benchmark}). This ensures the VLM being tested contains the relevant internal knowledge. 
    \item Remove an Image, QA pair if the VLM can correctly answer the QA pair when provided the \textit{visual reference} question and a trivial image. This filters questions that are trivially easy to answer using language priors from the sentence. 
\end{enumerate}

\begin{table*}[ht]
\centering
\begin{tabular}{@{}lllr@{}}
\toprule
\multicolumn{1}{c}{\textbf{Dataset}} & \multicolumn{1}{c}{\textbf{VLM}} & \multicolumn{1}{c}{\textbf{LM}} & \multicolumn{1}{l}{Datapoints} \\ \midrule
MNIST                                & LLaVA                            & Mistral-7B                      & 633                            \\
                                     &                                  & Vicuna-7B                       & 674                            \\
                                     &                                  & Vicuna-13B                      & 590                            \\
                                     & InstructBLIP                     & Vicuna-7B                       & 423                            \\
                                     &                                  & Vicuna-13B                      & 464                            \\
                                     & GPT4V                            & 4o-mini                         & 873                            \\
                                     &                                  & 4o                              & 983                            \\ \midrule
CIFAR100                             & LLaVA                            & Mistral-7B                      & 1411                           \\
                                     &                                  & Vicuna-7B                       & 1232                           \\
                                     &                                  & Vicuna-13B                      & 1222                           \\
                                     & InstructBLIP                     & Vicuna-7B                       & 73                             \\
                                     &                                  & Vicuna-13B                      & 155                            \\
                                     & GPT4V                            & 4o-mini                         & 1556                           \\
                                     &                                  & 4o                              & 1536                           \\ \midrule
Food101                              & LLaVA                            & Mistral-7B                      & 819                            \\
                                     &                                  & Vicuna-7B                       & 542                            \\
                                     &                                  & Vicuna-13B                      & 446                            \\
                                     & InstructBLIP                     & Vicuna-7B                       & 190                            \\
                                     &                                  & Vicuna-13B                      & 43                             \\
                                     & GPT4V                            & 4o-mini                         & 1387                              \\
                                     &                                  & 4o                              & 1296                              \\ \midrule
Landmarks                            & LLaVA                            & Mistral-7B                      & 359                            \\
                                     &                                  & Vicuna-7B                       & 643                            \\
                                     &                                  & Vicuna-13B                      & 689                            \\
                                     & InstructBLIP                     & Vicuna-7B                       & 0                              \\
                                     &                                  & Vicuna-13B                      & 121                            \\
                                     & GPT4V                            & 4o-mini                         & 810                              \\
                                     &                                  & 4o                              & 728                              \\ \midrule 
Average                            &                                  &                                 & 791.28                    \\ \bottomrule
\end{tabular}
\caption{Number of datapoints remaining after filtering}
\label{table:datapoints}
\end{table*}
After all steps of data filtering, the number of points in our testbed is dataset and VLM specific, with an average of 955 datapoints for each VLM and dataset combination. The exact breakdown is provided in Table~\ref{table:datapoints}. The main paper does not discuss results for VLM, dataset combinations that have fewer than 100 datapoints remaining after all filters. 

Finally, we conduct a round of human annotation. Three authors of this paper manually annotated a subset of 33 questions per dataset in CIFAR100, Food101 and Landmarks (a total of 99 questions). The questions were judged on two metrics: whether the question is relevant to the entity in the image, and whether the answer is a valid and correct answer to the provided question. 

Specifically, the following instructions were provided to each annotator:

\begin{verbatim}
You are to look at the following entity, 
QA-pair sets and annotate them for two criteria: 

1. Question Relevance
2. QA Pair Correctness

The standard for whether a question is relevant 
to an entity is whether or not the knowledge 
tested is associated with the stated entity 
(as opposed to some other entity).

The standard for Correctness is whether or not 
the answer provided an acceptable answer for 
the given question. The answer does not have 
to enumerate all possible answers to the 
question, but rather state one such answer 
to the question at hand. 
\end{verbatim}

\begin{table*}[t]
\begin{tabular}{@{}lrrrlrrrl@{}}\toprule
\multicolumn{4}{r}{Question Relevant to Entity}                                                          & \multicolumn{4}{r}{Answer Correct}                                                                       \\
\cmidrule(lr){2-5} \cmidrule(lr){6-9}
Dataset   & \multicolumn{1}{c}{Mean} & \multicolumn{1}{l}{Tot. Agree} & \multicolumn{1}{l}{Kappa} & Agreement   & \multicolumn{1}{c}{Mean} & \multicolumn{1}{l}{Tot. Agree} & \multicolumn{1}{l}{Kappa} & Agreement   \\ \hline
CIFAR100  & 86.86                    & 79.7                                & 12.85                     & Slight      & 92.73                    & 75.75                               & 17.23                     & Slight      \\
Food101   & 86.87                    & 90.9                                & 74.31                     & Substantial & 88.89                    & 81.81                               & 38.28                     & Fair        \\
Landmarks & 94.95                    & 93.94                               & 65.07                     & Substantial & 90.91                    & 90.91                               & 64.78                     & Substantial \\ \midrule
All       & 94.94                    & 93.94                               & 65.07                     & Substantial & 90.91                    & 90.91                               & 65.78                     & Substantial \\\bottomrule
\end{tabular}
\caption{Annotation results, judging whether the questions are relevant to the entities in the image, and whether the answers to the questions were correct. Total Agreement is the percentage of instances where all three annotators agreed. Kappa is the Fleiss Kappa score. The results show that the questions are relevant and correct with substantial agreement between annotators, suggesting that the questions are of high quality.}
\label{table:annotation}
\end{table*}

The results show (Table~\ref{table:annotation}) considerable agreement on most datasets, suggesting that the questions are of high quality. 

The prompts for all steps involving use of a LM are provided in Appendix~\ref{sec:appendix_prompts} and example QA pairs are given in Appendix~\ref{sec:appendix_examples}. 

\subsection{Prompts Used:}
\label{sec:appendix_prompts}
\noindent\textbf{QA Extraction:}
\begin{verbatim}
    You are a logical system tasked with 
    extracting questions for an entity 
    from a given text. The question 
    should have a unique answer, and 
    should be very short.

    Entity: Tench
    Text: The tench or doctor fish 
    (Tinca tinca) is a fresh- and 
    brackish-water fish of the order 
    Cypriniformes found throughout 
    Eurasia from Western Europe including 
    Britain and Ireland east into Asia 
    as far as the Ob and Yenisei Rivers. 
    It is also found in Lake Baikal. 
    It normally inhabits slow-moving 
    freshwater habitats, particularly 
    lakes and lowland rivers.

    Rationale: The tench is said to 
    also be called the doctor fish. 
    There is no other alternate name. 
    Question: What is another name 
    for the tench?
    Answer: doctor fish
    [SEP]
    Rationale: The order of the tench 
    is Cypriniformes. 
    Animals can only belong to one order.
    Question: What is the order 
    of the tench?
    Answer: Cypriniformes
    [SEP]
    Rationale: The usual habitat 
    of the tench is freshwater. 
    Question: What kind of water does 
    the tench usually live in?
    Answer: freshwater 
    [STOP]

    Entity: Baklava
    Text: Baklava is a layered pastry 
    dessert made of filo pastry, 
    filled with chopped nuts, 
    and sweetened with syrup or honey.
    It was one of the most popular 
    sweet pastries of Ottoman cuisine.
    There are several theories for 
    the origin of the pre-Ottoman 
    version of the dish. 
    In modern times, it is a common 
    dessert among cuisines of 
    countries in West Asia, 
    Southeast Europe, Central Asia, 
    and North Africa. It is also 
    enjoyed in Pakistan and 
    Afghanistan, where, although 
    not a traditional sweet, it 
    has carved out a niche in 
    urban centers.
    Rationale: Baklava is 
    associated with Ottoman cuisine.
    Question: What ancient 
    cuisine is Baklava associated with?
    Answer: Ottoman
    [SEP]
    Rationale: Baklava is made of 
    filo pastry.
    Question: What kind of pastry 
    is Baklava made of?
    Answer: filo
    [STOP]

    Entity: <NEW ENTITY>
    Text: <SPLIT FROM WIKIPEDIA>
\end{verbatim}

\noindent\textbf{Identifying whether a question is ambiguous}:

\begin{verbatim}
    Given a text and a question, 
    judge whether the question has 
    a unique answer, or can be 
    answered with multiple valid responses.

    Text: The tench or doctor ...
    Question: What is the tench 
    also known as?
    Rationale: The text mentions only 
    one other name for the tench, 
    which is doctor fish.
    Judgment: Unique [STOP]

    Text: The tench ...
    Question: Which lake is 
    the tench found in? 

    Rationale: The text mentions 
    that the tench is found in 
    Lake Baikal. However, it is 
    very likely that the tench 
    is found in other lakes as well.
    Judgment: Multiple [STOP]

    Text: <SPLIT FROM WIKIPEDIA>
    Question: <GENERATED QUESTION>
\end{verbatim}

\noindent\textbf{Question Answering:}
\begin{verbatim}
    You are a knowledgeable 
    system tasked with answering 
    the question based on your knowledge.

    Question: What is the tench 
    also known as?
    Answer: doctor fish [STOP]

    Question: What kind of pastry 
    is Baklava made of?
    Answer: Filo [STOP]

    Question: In which city is 
    the Eiffel Tower located?
    Answer: Paris [STOP]

    Question: <GENERATED QUESTION>    
\end{verbatim}

\noindent\textbf{Duplicate Identification:}
\begin{verbatim}
    You are a logical system tasked 
    with determining if two question
    answer pairs are duplicates of 
    each other.

    Question: What is the tench 
    also known as?
    Answer: doctor fish
    Question: What is another name 
    for the tench?
    Answer: the doctor fish
    Rationale: The two questions 
    are asking the same thing, and 
    have the same answer.
    Judgment: Duplicate [STOP]

    Question: What pastry is 
    Baklava made of?
    Answer: filo
    Question: What kind of nuts 
    are used to make Baklava?
    Answer: walnuts
    Rationale: The two questions 
    are asking about different 
    things, and have 
    different answers.
    Judgment: Unique [STOP]

    Question: <GENERATED Q1>
    Answer: <GENERATED A1>
    Question: <GENERATED Q2>
    Answer: <GENERATED A2>
\end{verbatim}

\noindent\textbf{MCQA Conversion}:
\begin{verbatim}
    You are a logical system tasked 
    with generating incorrect options 
    for multiple choice questions. 
    You are given the text, question 
    and answer. Come up with a numbered 
    list of three plausible 
    but incorrect options

    Text: The tench ...
    Question: What is the tench 
    also known as?
    Correct Answer: doctor fish
    Incorrect Option 1: miracle fish
    Incorrect Option 2: salmon 
    Incorrect Option 3: hidden fish
    [STOP]

    Text: <TEXT>
    Question: <QUESTION>
    Correct Answer: <CORRECT ANSWER>
\end{verbatim}

\subsection{QA Pair Examples:}
\label{sec:appendix_examples}
The following QA pairs are randomly sampled from their datasets:

\noindent\textbf{CIFAR100}:
\begin{verbatim}
    Entity: Television
    Question: What is the title of a book 
    about the invention of the television?
    Answer: Tube: the Invention of Television

    Entity: road
    Question: Which road reaches the North 
    Slope of Alaska?
    Answer: Dalton Highway

    Entity: sunflower
    Question: When were sunflower seeds 
    brought to Europe?
    Answer: 16th century

    Entity: snake
    Question: What Hindu festival is 
    associated with snakes?
    Answer: Nag Panchami
\end{verbatim}

\noindent\textbf{Food101}:
\begin{verbatim}
    Entity: oysters
    Question: What sauce is often paired 
    with oysters?
    Answer: mignonette

    Entity: Edamame
    Question: In which cuisine is 
    edamame a common side dish?
    Answer: Japanese

    Entity: sushi
    Question: How is sushi served in the 
    kaiten zushi style?
    Answer: on a conveyor belt

    Entity: Ice cream
    Question: Can ice cream be made 
    with a blender?
    Answer: Yes
\end{verbatim}

\noindent\textbf{Landmarks}:
\begin{verbatim}
    Entity: Grand Canyon
    Question: Who led an expedition to the 
    Grand Canyon?
    Answer: John Wesley Powell

    Entity: Gateway of India
    Question: In which city is the Gateway of 
    India located?
    Answer: Mumbai

    Entity: Eiffel Tower
    Question: When did the Eiffel Tower 
    stop broadcasting analogue 
    television signals?
    Answer: 2011

    Entity: Machu Pichu
    Question: In which country is Machu 
    Picchu located?
    Answer: Peru

    Entity: Golden Gate Bridge
    Question: What speed do jumpers hit the 
    water at when jumping from the 
    Golden Gate Bridge?
    Answer: 75 mph
\end{verbatim}

\section{Vision Language Model Inference}
\label{sec:appendix_vlm_inference}
All VLM inference uses greedy decoding and is hence deterministic. 

To measure the performance of a VLM with trivial images, we use four different kinds of trivial images:
\begin{enumerate}
    \item Black: A pure black image
    \item White: A pure white image
    \item Noise: A noised image where each colour channel (RGB) value for each pixel is sampled uniformly from 0 to 255.  
    \item None: For VLMs that leave the LM unchanged (e.g. InstructBLIP), we do only a forward pass through the LM. For VLMs which tune the LM in any way (e.g. LLaVA), we pass in a null image. The image representation that results is then an average image representation over all images seen during the VLMs training. 
\end{enumerate}

We collect the output of the VLM from all four of these trivial images (with the same text prompt). The final output of the VLM in this setting is the result of a majority vote on the output (or the modal output from all trivial images). Ties are broken randomly.  

\section{Robustness to choice of metric}
\label{sec:appendix_metric}

\begin{figure*}[t]
    \centering 

    \begin{subfigure}[t]{0.49\textwidth} 
        \centering
        \includegraphics[width=0.99\linewidth]{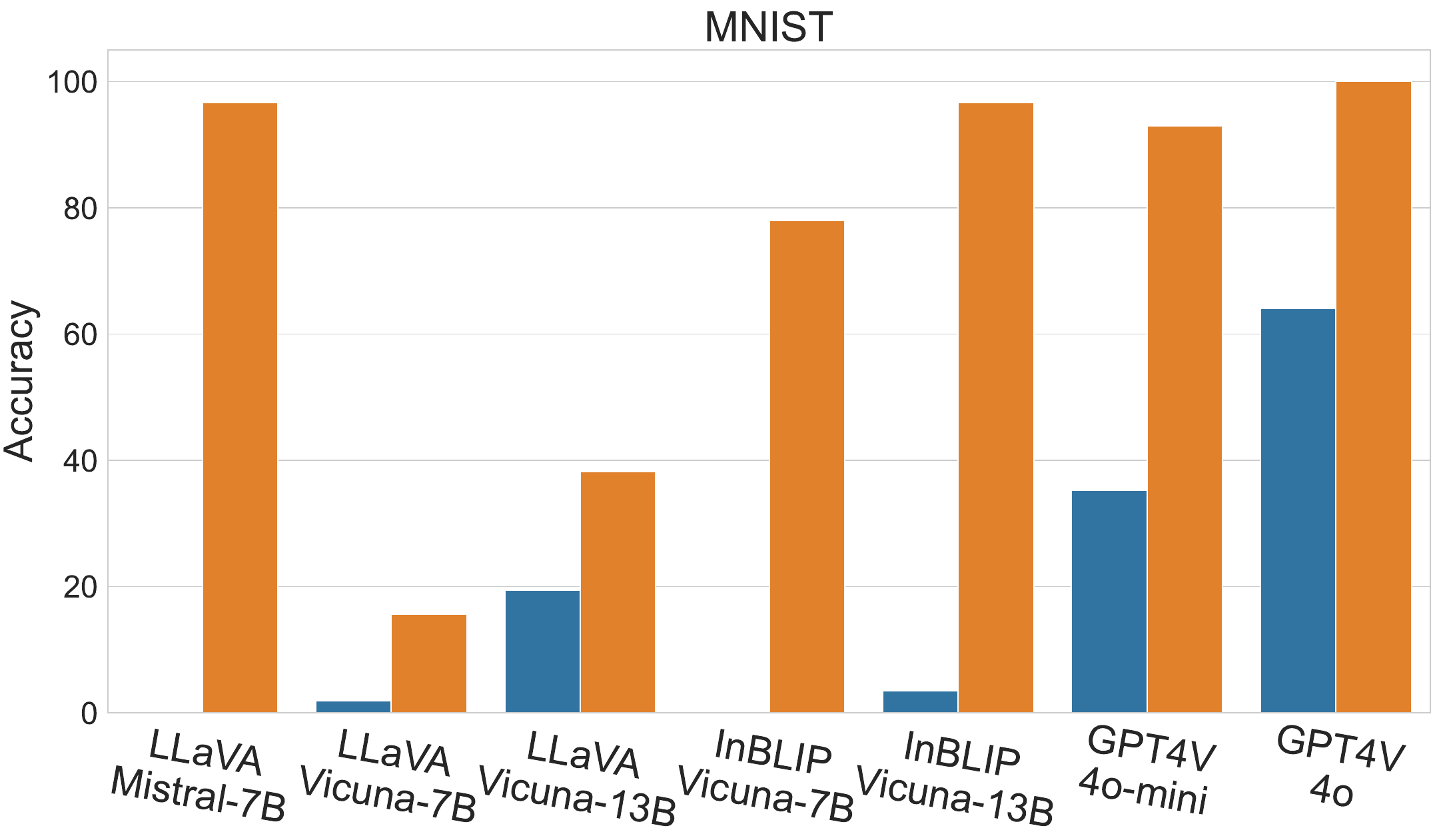}
    \end{subfigure}
    \hspace{0.1cm}
    \begin{subfigure}[t]{0.49\textwidth}
        \centering
        \includegraphics[width=0.99\linewidth]{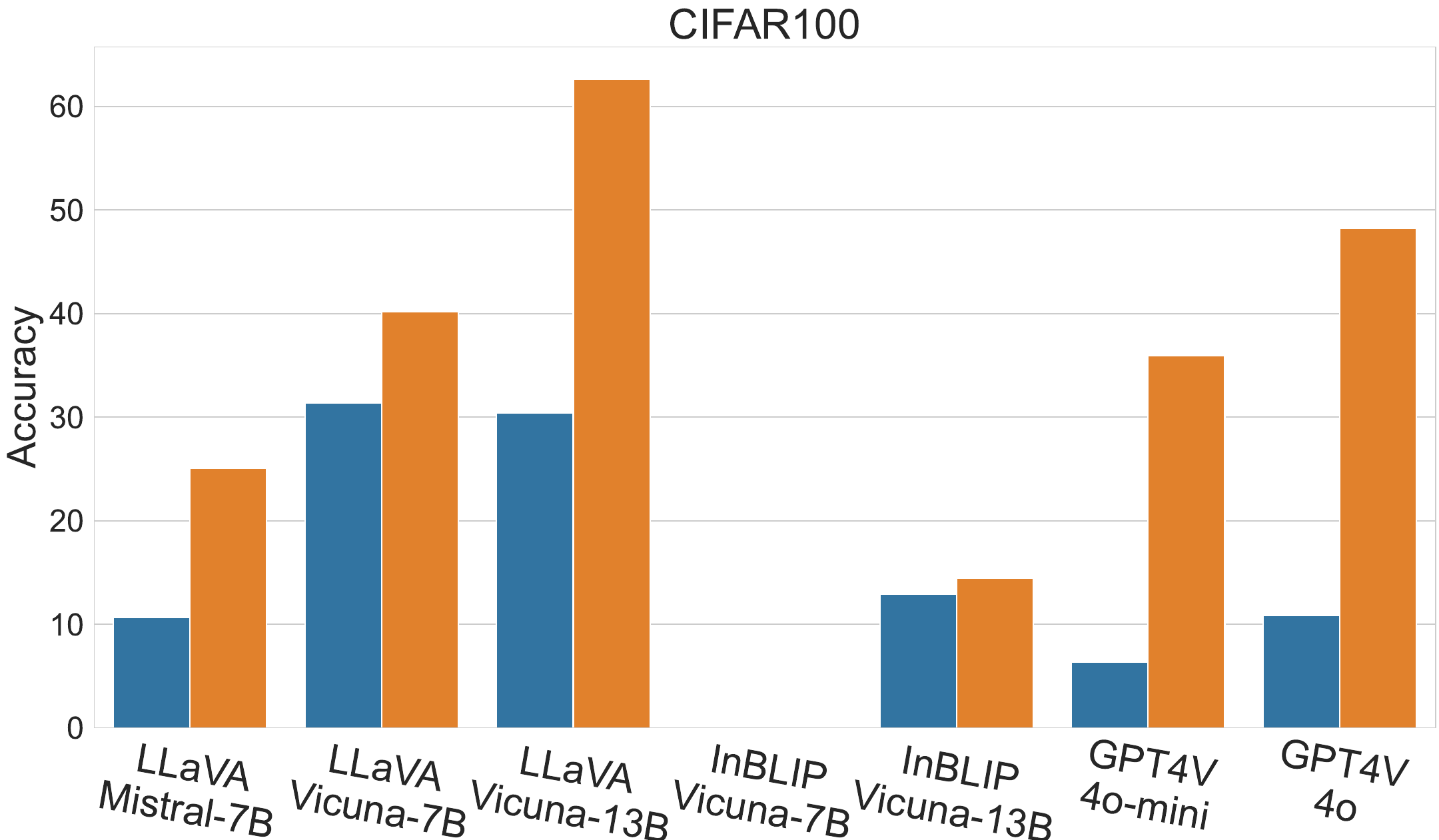}
    \end{subfigure}%

    \vspace{0.5cm} 

    \begin{subfigure}[t]{0.49\textwidth}
        \centering
        \includegraphics[width=0.99\linewidth]{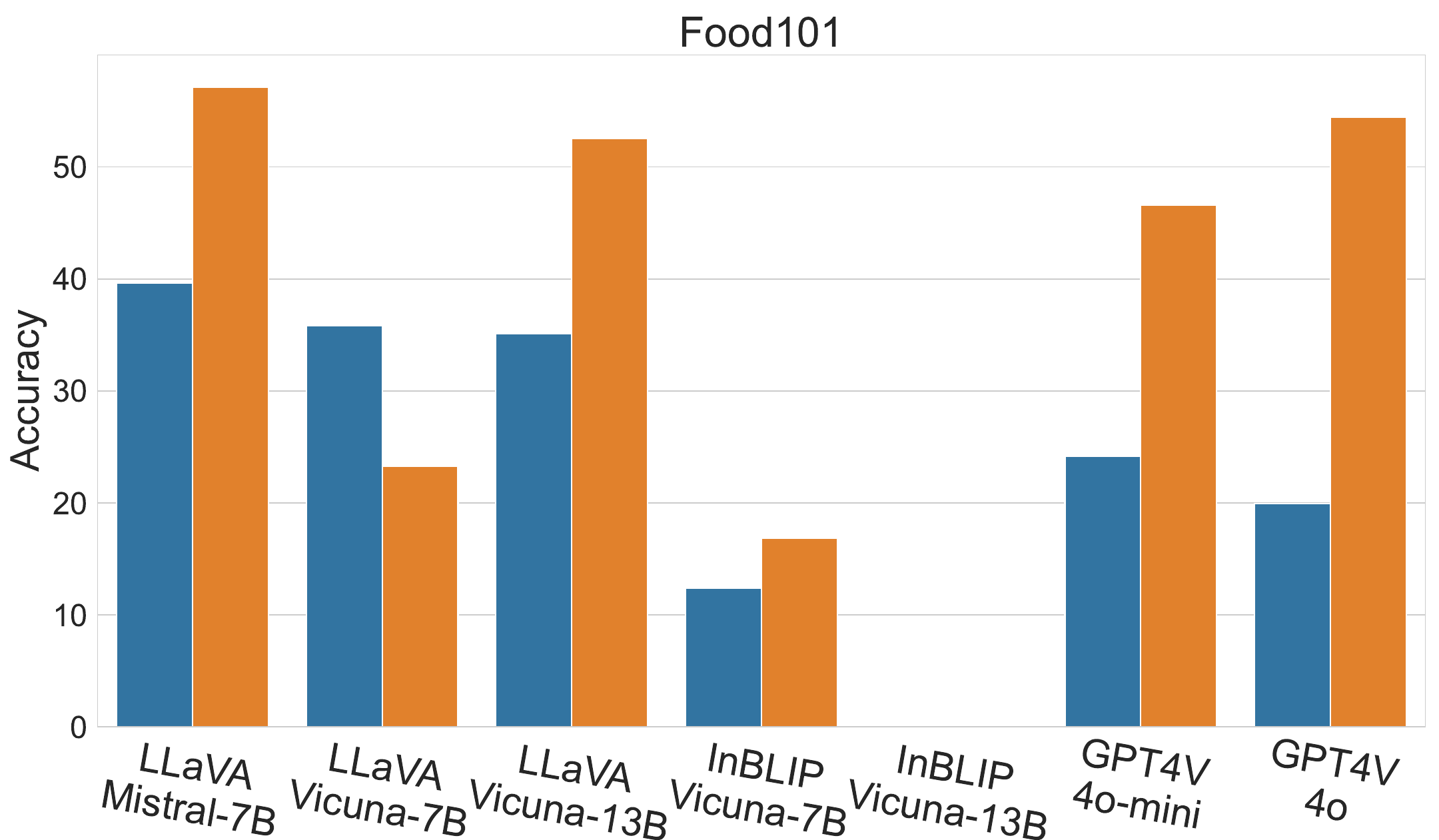}
    \end{subfigure}%
    \hspace{0.1cm} 
    \begin{subfigure}[t]{0.49\textwidth}
        \centering
        \includegraphics[width=0.99\linewidth]{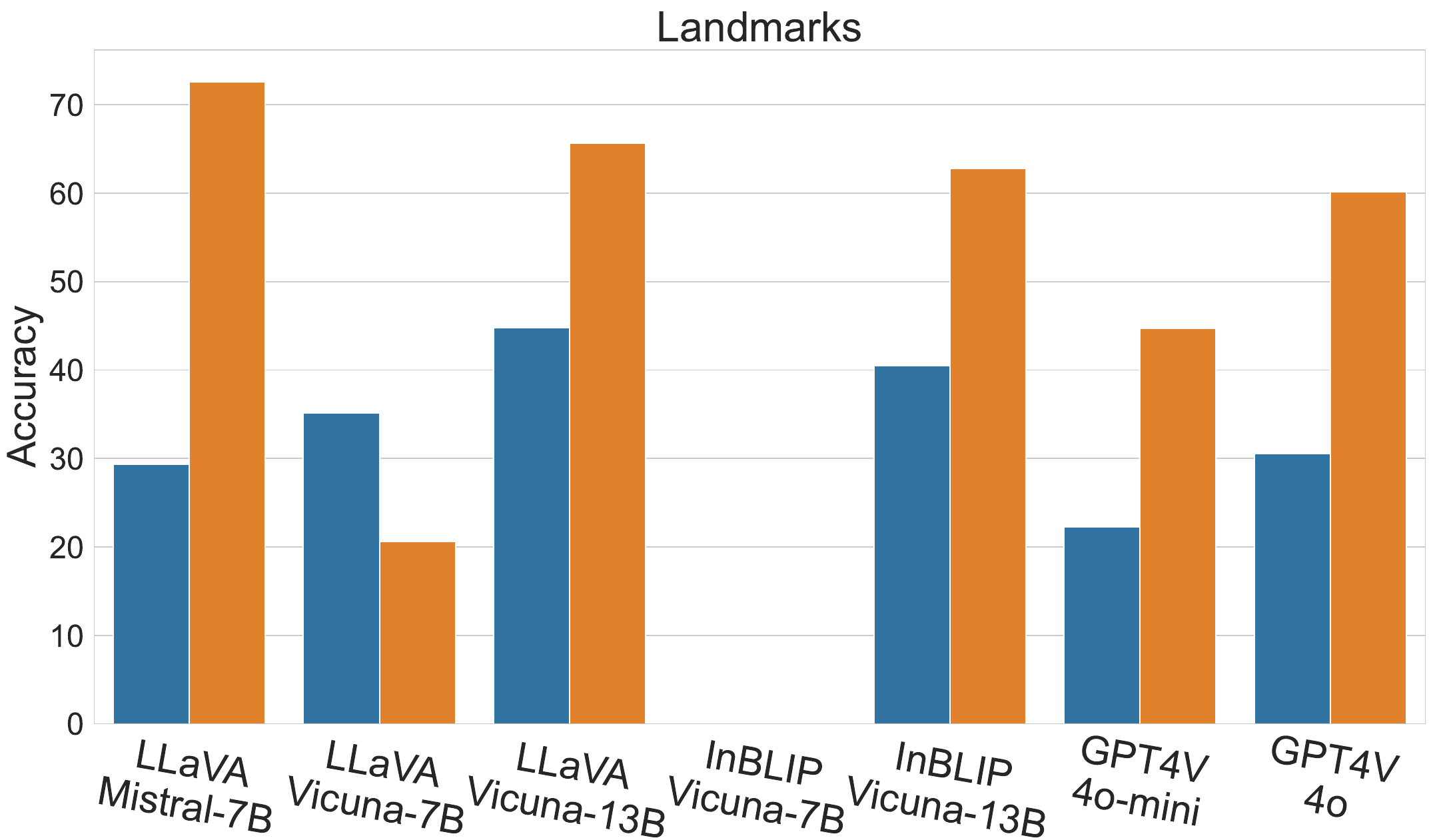}
    \end{subfigure}%
    \caption{Exact Match Performance on VLMs when forced to rely on visual representations (\textcolor{blue}{Visual}) as well as textual representations (\textcolor{orange}{Text Only}) of an entity.}
    \label{fig:exactmatch}

\end{figure*}

\begin{figure*}[t]
    \centering 

    \begin{subfigure}[t]{0.49\textwidth} 
        \centering
        \includegraphics[width=0.99\linewidth]{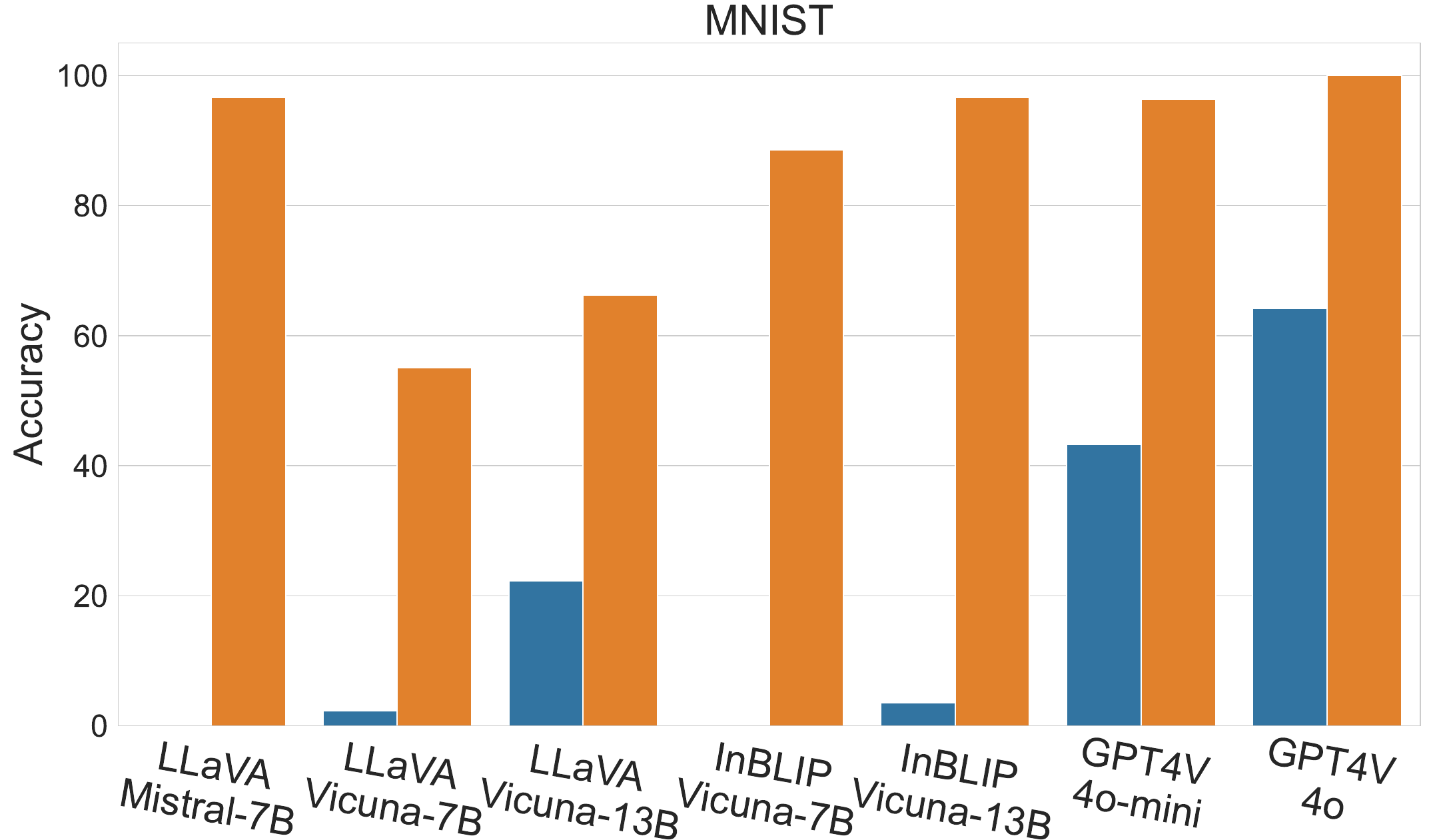}
    \end{subfigure}
    \hspace{0.1cm}
    \begin{subfigure}[t]{0.49\textwidth}
        \centering
        \includegraphics[width=0.99\linewidth]{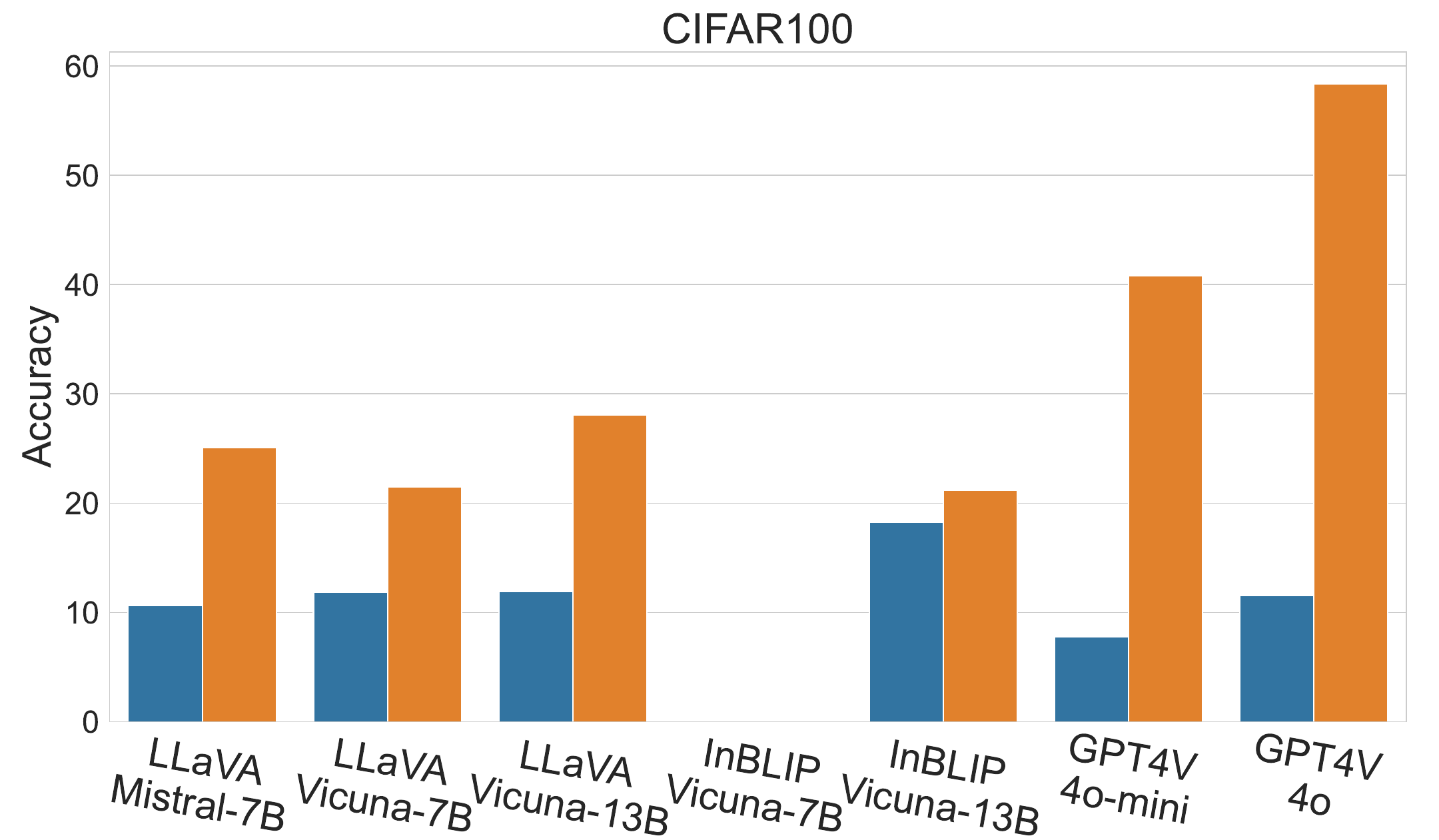}
    \end{subfigure}%

    \vspace{0.5cm} 

    \begin{subfigure}[t]{0.49\textwidth}
        \centering
        \includegraphics[width=0.99\linewidth]{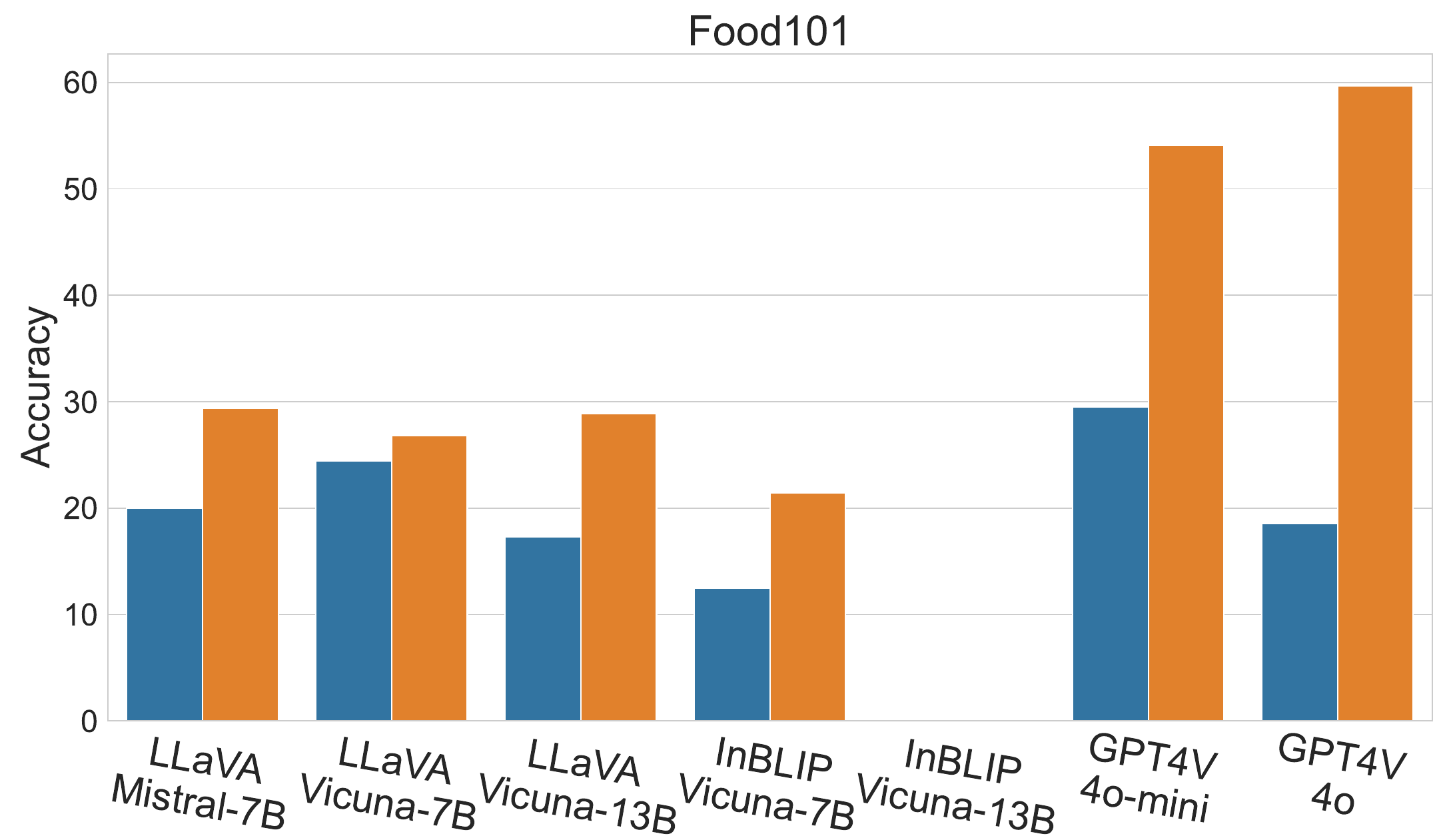}
    \end{subfigure}%
    \hspace{0.1cm} 
    \begin{subfigure}[t]{0.49\textwidth}
        \centering
        \includegraphics[width=0.99\linewidth]{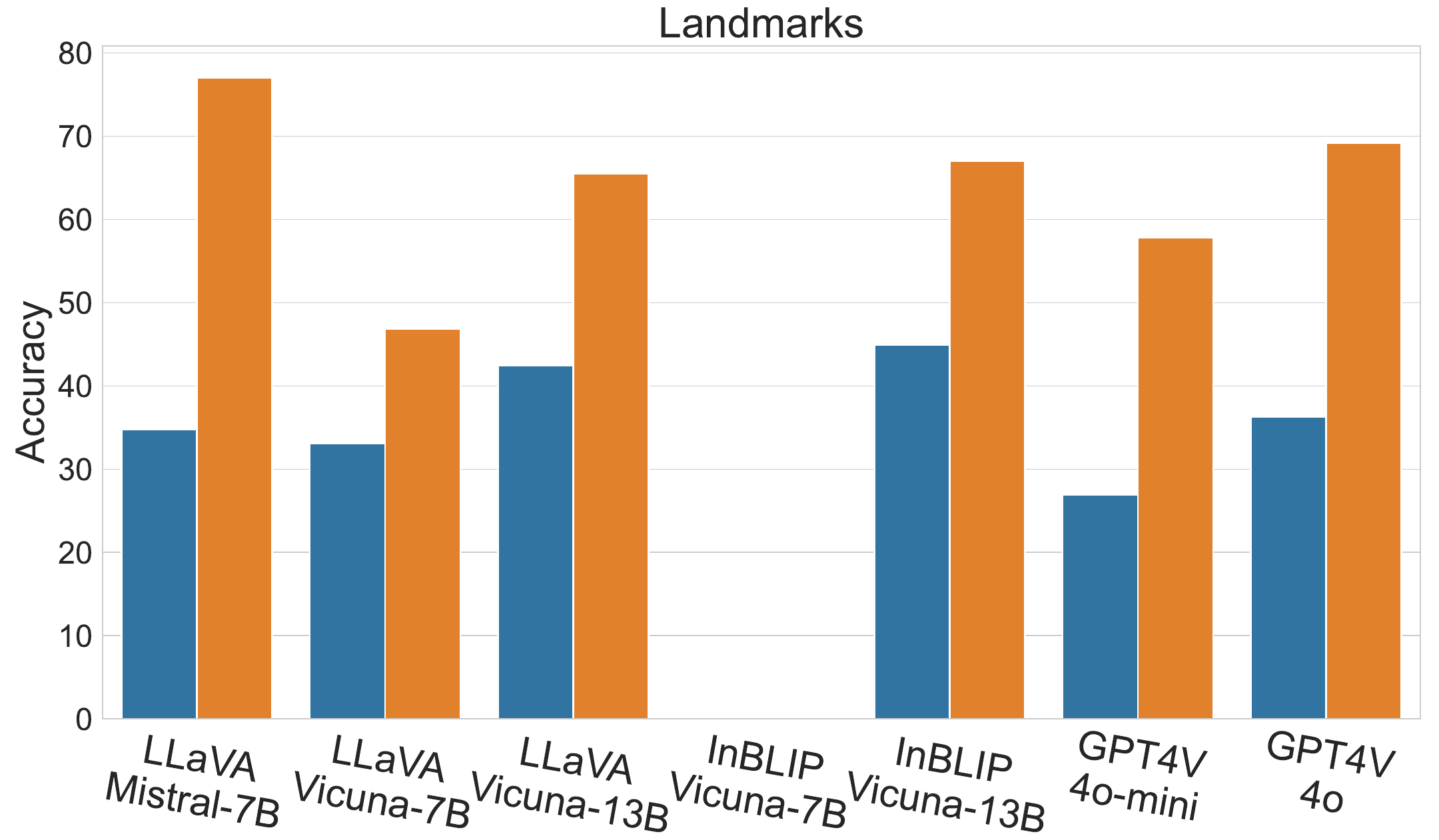}
    \end{subfigure}%
    \caption{BLEU Performance on VLMs when forced to rely on visual representations (\textcolor{blue}{Visual}) as well as textual representations (\textcolor{orange}{Text Only}) of an entity.}
    \label{fig:bleu}

\end{figure*}

The main text shows the results for the two way inclusion metric. Here we show results for exact match (Figure~\ref{fig:exactmatch}) and BLEU (Figure~\ref{fig:bleu}). In all cases, we see a significant performance gap between the TextOnly and the Visual Setting

\section{Hardware for experiments}
\label{sec:appendix_hardware}
The experiments conducted in this paper were run on three distinct compute clusters, with different specifications for each one. We list all of the core details below:
\begin{enumerate}
    \item A cluster with 5 RTX 2080 TI GPUs. 64GB of RAM and 5 Intel(R) Xeon(R) Gold 5215 CPUs @2.50 GHz
    \item A cluster with 8 NVIDIA L40s GPUs (48G GPU memory each). 800GB of RAM and 256 AMD EPYC 9554 64-Core Processor.
    \item A cluster with 4 NVIDIA A100 GPUs (40G or 80G GPU memory each), 514GB of RAM, and 96-core Intel(R) Xeon(R) Platinum 8272CL CPUs @2.60GHz.
\end{enumerate}

The time taken to run each part of the experiment varies with the hardware used. However typically all experiments for a single dataset (start to finish from data generation to final results) can be completed within 4 GPU days.

\end{document}